\documentclass[lettersize,onecolumn]{IEEEtran}
\usepackage{amsmath,amsfonts}
\usepackage{amssymb}
\usepackage{amsthm}

\usepackage{algorithmic}
\usepackage{algorithm}
\usepackage{array}
\usepackage[caption=false,font=normalsize,labelfont=sf,textfont=sf]{subfig}
\usepackage{textcomp}
\usepackage{stfloats}
\usepackage{url}
\usepackage{verbatim}
\usepackage{graphicx}
\usepackage{tikz}
\usetikzlibrary{calc}
\usepackage{cite}
\usepackage{bm}
\newtheorem{lemma}{Lemma}
\newtheorem{theorem}{Theorem}
\newtheorem{proposition}{Proposition}
\newtheoremstyle{italicremark}
  {0pt}           % Space above (set to 0)
  {0pt}           % Space below (set to 0)
  {\normalfont}   % Body text font (Normal/Roman)
  {}              % Indent amount
  {\itshape}      % Label font (Italic "Remark")
  {.}             % Punctuation after "Remark 1"
  {.5em}          % Space after the punctuation
  {}              % Head spec

\theoremstyle{italicremark}
\newtheorem{remark}{Remark}

\begin{document}

\title{Effect of Spherical Parameter Space on Number of Non-Forgetting Tasks and Capacity of Continual Learning}

\author{Nazanin Mirhosseini \\ \textit{Colorado State University} \\ Fort Collins, Co, USA \\ nazmir@colostate.edu}

\maketitle

\begin{abstract}

	We propose a continuous learning formulation for $T$ non-forgetting tasks with the spherical parameter space $\mathbb{S}^{d-1}(\sqrt{d})$ in asymptotic regime $d\rightarrow \infty$. Each task $i$ has an optimum $\bm{\Theta}_i$, minimizing its corresponding loss function $R_i(\bm{w})$ on sphere. We reduce each task $i$ to an acceptable loss region modeled by a geodesic ball $B_{\bm{\Theta}_i}(r_i(d))$ on $\mathbb{S}^{d-1}(\sqrt{d})$ centered at $\bm{\Theta}_i$ with radius $r_i(d)$. To model task dependencies (similarities), task optima $\bm{\Theta}_i$ are distributed by a Neyman-Scott clustered point process $\Phi$. Cluster centers $\{\bm{Y}_j\}_{j=1}^{N_p}$ form a homogeneous Poisson point process on sphere, while intra-cluster task optima follow a von Mises-Fisher distribution with mean $\bm{Y}_j$ and concentration $\kappa$. Each task $t\in \{1, ..., T\}$ learning algorithm is considered to be a channel $Q_{t}(\mathrm{d}\bm{w}_t | \bm{w}_{t-1}, \mathcal{K}_{t-1}(\bm{\theta}_{1:t-1}), \bm{\theta}_t)$, where $\bm{W}_t$ is the learned parameter of task $t$ and $\mathcal{K}_{t-1}(\bm{\Theta}_{1:t-1})=\bigcap_{i=1}^{t-1}B_{\bm{\Theta}_i}(r_i(d))$. Since in continual learning, the learner does not need to reach $\bm{\Theta}_t$ exactly to successfully learn task $t$, to avoid forgetting as well as reliable learning, it only needs for the learned parameter vector $\bm{W}_{t}$ to enter $B_{\bm{\Theta}_t}(r_t(d)) \cap \mathcal{K}_{t-1}(\bm{\Theta}_{1:t-1})$. Using this model, we derive an upper bound on $T$ ensuring $\mathbb{P}[\mathcal{K}_{T}(\bm{\Theta}_{1:T}) \neq \emptyset] \rightarrow 1$ as $d\rightarrow \infty$ and $\kappa/d \rightarrow \alpha$, where $\alpha>0$ is a constant. Finally, for $T$ tasks sampled uniformly from $\Phi$, we derive an upper bound for total information theoretic capacity under the non-forgetting constraint for each task.

\end{abstract}

\section{Introduction}
Continual learning studies how a model can learn from a sequence of evolving tasks or data distributions by incrementally updating its knowledge over time, rather than optimizing once for a static objective \cite{Wang2024}, \cite{DeLange2023}. A key challenge is that, when transitioning to a new task, the learner often has little or no access to data from previous training stages. Consequently, gradient updates optimized for the current data distribution can interfere with or overwrite representations that supported earlier tasks, resulting in catastrophic forgetting \cite{Wang2024}, \cite{Shi2025}. Accordingly, continual learning is commonly framed as a \textit{stability-plasticity} problem: the learner must remain sufficiently plastic to acquire new information while remaining stable enough to preserve previously acquired knowledge \cite{Wang2024, DeLange2023, Mirzadeh2020, Parisi2019}.

Catastrophic forgetting in continual learning is not merely an optimization failure, but a fundamental structural consequence of sequential learning within a shared parameterization \cite{Wang2024}. As a learner accumulates tasks, each new objective must be learned within a parameter region that preserves performance on previous data, causing the joint feasible set to become progressively narrower and more irregular \cite{Wang2024, Farajtabar2020, Mirzadeh2021}. Recent studies emphasize that this phenomenon is deeply governed by the geometry of the loss landscape: forgetting is shaped by the eigenspectrum of the Hessian matrix, the length of the optimization path, and the sharpness of the local minima \cite{Wang2024, Mirzadeh2020, Mirzadeh2021, Yang2025}. Specifically, the maximum eigenvalue of the Hessian quantifies the most fragile curvature direction around an old task solution such that when this curvature is steep, even minor parameter updates along sensitive directions cause substantial degradation \cite{Kong2024}. Consequently, mitigating forgetting requires navigation of parameter space, whether by projecting new gradients orthogonally to previous task subspaces \cite{Farajtabar2020}, managing the principal angles between gradient subspaces \cite{Steele2026}, or accounting for the inherent similarities and sequence order of the tasks themselves \cite{Lin2023}. These observations also suggest that the central question can be viewed geometrically: whether a parameter configuration exists that remains simultaneously viable for all tasks encountered so far. Finding a shared parameter vector across incremental tasks corresponds to solving a set intersection decision problem over viable parameter spaces \cite{wolczyk2022, Masip2026}. Because viable parameter regions have arbitrary, non-convex shapes, identifying an optimal shared solution is generally NP-hard \cite{Knoblauch2020}. To derive computationally tractable guarantees, these feasible sets are approximated as intersecting hyperrectangles using interval arithmetic, reducing optimization complexity back to polynomial time \cite{wolczyk2022}. While this approximation provides an important route toward tractable analysis, reducing computational complexity is not the primary objective of the present work. Instead, we focus on the existence of a nonempty shared feasible region and, in particular, on how the geometry and curvature associated with individual tasks determine the number of sequential tasks that can be accommodated before this region becomes empty.

Building upon these geometric and curvature-based insights, this paper analyzes a mathematical framework that formalizes the continual learning parameter space as a hypersphere. 
Here, the hypersphere should be understood as a controlled geometric abstraction that provides a bounded and analytically tractable parameter domain, rather than as a claim that the actual parameter space of a neural network is intrinsically spherical. This choice allows parameter differences to be represented through angular or geodesic distances while providing a common boundary on which task optima can be placed, making it possible to study overlap of feasible regions in a geometrically consistent manner. In our formulation, each task optimum is represented as a specific point on the spherical surface, acting as the center of a geodesic ball. The radius of this ball represents the acceptable tolerance loss for parameter updates. The nonempty intersection of these successive geodesic balls constitutes the feasible region for the learner. By mapping the parameter space to this topology, we establish a theoretical link between the local curvature and the continual learner capacity. We prove that the maximum number of sequential tasks a learner can learn, which is the threshold at which the feasible region becomes empty, depends on the maximum eigenvalue of the task Hessians. Furthermore, because real-world continual learning is dependent on task relatedness \cite{Lin2023}, we model the underlying geometry of task similarity by distributing the task optima across the hypersphere using a Neyman-Scott point process. This allows us to theoretically capture the clustering behavior of related tasks \cite{Lawrence2016}, providing a geometric perspective that explains how task distribution and spherical loss curvature interact to determine the limits of continual learning.

From information theoretic perspective, recent work \cite{Cheng2026} formalizes continual learning as a constrained online coding problem and establishes an impossibility triangle, showing that sequential state-based learners cannot simultaneously achieve zero forgetting, online learning, and finite parameterization. We extend the information theoretic perspective on continual learning in the asymptotic regime with spherical parameter space. Starting from an initial parameter $\bm{W}_0$ distributed uniformly on the sphere, each newly learned task induces a transition to another point on the sphere. However, to ensure learning without forgetting, the updated parameter must remain within a feasible region determined by the intersection of previously learned tasks loss regions. We therefore view the learned parameters as a moving point on the sphere, initialized at $\bm{W}_0$, whose trajectory is induced by the sequence of task optima and constrained to remain within the feasible region. The channel capacity is then defined in terms of the mutual information between the task optima and the resulting learned parameters. This geometric and asymptotic formulation allows us to characterize the amount of information that can be stored in the evolving parameter state while maintaining reliability and avoiding forgetting.

This paper is organized as follows. In Section \ref{sec1}, we elaborate more on spherical assumption of parameter space in high dimension and show that the acceptable loss region for each task can be approximated by a geodesic ball on sphere centered at the task optima. In Section \ref{sec2}, we elaborate on the Neyman-Scott clustered point process $\Phi$ on the sphere as a distribution of task optima. In Section \ref{sec3}, we geometrically model the forgetting phenomenon and, as a result, derive a bound on the number of tasks that can be learned without forgetting. We further formulate the learning of a sequence of tasks as a sequence of channels, define the capacity of the continual learning and derive an upper bound on the capacity.

\textit{Notations}: We denote a vector using bold font, $\bm{v}$, and its $i$-th entry by $v_i$. To emphasize the dimension of a vector, we also denote a vector of dimension $d$ by $v_{1:d}$. Matrices are denoted by roman bold font $\mathbf{M}$. Convergence in probability is denoted by $\xrightarrow{P}$, and convergence in probability to $0$ is also denoted by $o_P(1)$. The uniform distribution on a set $\mathcal{A}$ is denoted by $\mathrm{Uniform}(\mathcal{A})$, and the Poisson distribution with intensity $c$ is denoted by $\mathrm{Poisson}(c)$. The complement of a set $\mathcal{A}$ is denoted by $\mathcal{A}^c$. The indicator function is denoted by $\mathbf{1}\{\cdot\}$. The inner product of vectors $\bm{v}, \bm{u} \in \mathbb{R}^d$ is denoted by $\langle \bm{v}, \bm{u} \rangle$. A $d\times d$ identity matrix is denoted by $\mathbf{I}_d$.
We denote the $2$-norm of a vector $\bm{v}$ as $\|\bm{v}\|$. The operator norm of a matrix $\mathbf{M}$ is represented by $\| \mathbf{M}\|_{\mathrm{op}}$. We denote the first standard basis vector as $\bm{e}_1= (1,0,...,0)$.

\section{Spherical Parameter Space and Acceptable Loss Regions}\label{sec1}
The structure of the parameter space directly governs the trade-off between retaining past knowledge and accommodating incoming tasks \cite{Wang2024}. Standard unconstrained Euclidean spaces lead to catastrophic interference because shared parameter coordinates drift uncontrolled across tasks \cite{He2026, Garg2026}. Moving from standard Euclidean configurations, constraining parameters to a spherical manifold provides distinct inductive biases for continual learning \cite{Ren2025, Gao2023}, as well as decouples the parameter vector's magnitude from its direction. Moreover, spherical spaces facilitate uniform angular regularization, ensuring that task prototypes and class representations remain maximally separable while leaving low-curvature directional to remain available for new learning \cite{Ren2025}.

Let $\mathbb{S}^{d-1}(\sqrt{d}) = \{ \mbox{\boldmath $w$}\in \mathbb{R}^d: \| \mbox{\boldmath $w$}\| = \sqrt{d}\}$ denote the  hypersphere in $\mathbb{R}^{d}$ with radius $\sqrt{d}$ equipped with the normalized Haar measure $\sigma$ such that $\sigma\left(\mathbb{S}^{d-1}(\sqrt{d})\right)=1$, representing the parameter space.
In standard neural network initialization schemes, the weights of each layer are often modeled as independent Gaussian random variables whose variance is scaled according to the layer dimensions \cite{Glorot2010, He2015}. To simplify the analysis, we consider the isotropic normalization in which the initial parameter vector is $\bm{W}_0\sim\mathcal N(\mathbf 0,\mathbf{I}_d)$.
This normalization can be viewed as a rescaled version of the layer dependent initialization and allows us to focus on the geometry of a high dimensional Gaussian vector. Although such a Gaussian vector has support on all of $\mathbb R^d$, its Euclidean norm is strongly concentrated around $\sqrt d$. In particular, for any fixed $\epsilon>0$, by concentration of measure \cite{Vershynin}, we have
\begin{eqnarray}
	\mathbb P\left[\left|\|\bm{W}_0\|-\sqrt d\right|>\epsilon\sqrt d\right] \leq 2\exp\{-c_\epsilon d\},
\end{eqnarray}
where $0<\epsilon<1$ and $c_\epsilon>0$. Thus, with probability at least $1-2\exp\{-c_\epsilon d\}$, a Gaussian task vector has its norm in the thin annulus $[(1-\epsilon)\sqrt d,(1+\epsilon)\sqrt d]$. Hence, in high dimensions, the Gaussian measure is concentrated near the sphere of radius $\sqrt d$. Equivalently, replacing $\mathbb{R}^d$ by $\mathbb{S}^{d-1}(\sqrt{d})$ amounts to neglecting the radial fluctuations of the Gaussian distribution, while retaining its isotropic angular component.

Let task $i$ be presented by its optimal feature alignment vector $\bm{\Theta}_{i}\in \mathbb{S}^{d-1}(\sqrt{d})$ such that 
\begin{equation}
	R_i(\bm{\Theta}_{i}) = \min_{\bm{w}\in \mathbb{S}^{d-1}(\sqrt{d})} R_i(\bm{w}),
\end{equation}
where $R_i(\bm{w})$ is the loss function corresponding to task $i$. To make derivations tractable in this paper, we assume that for every task $i$, the loss function $R_i$ has a local minimum on $\mathbb{S}^{d-1}(\sqrt{d})$. For a fixed $\eta \geq 0$ tolerated loss increase per dimension, the acceptable loss region for task $i$ is defined as
\begin{eqnarray}
	\mathcal{L}_i := \{ \bm{w}\in \mathbb{S}^{d-1}(\sqrt{d}): R_i(\bm{w}) - R_i(\bm{\Theta}_i) \leq d \eta\}.\label{Loss}
\end{eqnarray}
We denote the geodesic distance between two points $\mbox{\boldmath $u$}, \mbox{\boldmath $v$} \in \mathbb{S}^{d-1}(\sqrt{d})$ by
\begin{eqnarray}
	d_{geo}(\mbox{\boldmath $u$},\mbox{\boldmath $v$}) : = \sqrt{d}\arccos \left(\frac{\langle \bm{u},\bm{v}\rangle}{d}\right). \label{geodis}
\end{eqnarray}
Using the geodesic distance, the geodesic ball centered at $\bm{\Theta}_i$ with radius $r_i(d)$ is denoted by 
\begin{eqnarray}
 	B_{\bm{\Theta}_i}(r_i(d)): =\{\bm{w}\in \mathbb{S}^{d-1}(\sqrt{d}): d_{geo}(\bm{\Theta}_i,\bm{w}) \leq r_i(d)\}.
\end{eqnarray}  

\begin{lemma}\label{Lemma1}
Let $\mathcal{T}_{\bm{\Theta}_i}=\left\{\bm{v}\in\mathbb{R}^{d}: \langle \bm{v},\bm{\Theta}_i\rangle=0\right\}$ be the tangent space to $\mathbb{S}^{d-1}(\sqrt{d})$ at $\bm{\Theta}_i$. Let $f_i(\bm v)$ be the loss function $R_i(\bm w)$ expressed in the local geodesic coordinates around $\bm{\Theta}_i$. Assume that $\bm{\Theta}_i$ is a local minimizer of $R_i$, and that the Hessian matrix $\mathbf H_i$ of $f_i$ at $\bm 0$ is isotropic on $\mathcal{T}_{\bm{\Theta}_i}$, i.e., $\mathbf H_i= \nabla^2 f_i(\bm 0)=\lambda_i\mathbf I_{d-1}$, for
$\lambda_i>0$.
Assume that, for every fixed $0<C<\pi$,
\begin{eqnarray}
	\sup_{\|\bm z\|\leq C}
\left|\frac{1}{d}\left(f_i(\sqrt d\,\bm z)-f_i(\bm 0)\right)-\frac{\lambda_i}{2}\|\bm z\|^2\right|\longrightarrow 0, \qquad\text{as }d\to\infty, \label{UniformAssumption}
\end{eqnarray}
where $\bm z=\frac{\bm v}{\sqrt d}$ for all $\bm v\in\mathcal{T}_{\bm{\Theta}_i}$.
Then the acceptable loss region $\mathcal{L}_i$ is asymptotically $B_{\bm{\Theta}_i}(r_i(d))$ with $r_i(d)=\sqrt{{2\eta}/{\lambda_i}}\sqrt d\,(1+o(1))$.
\end{lemma}
\begin{proof}
	Proof is given in Appendix \ref{ProofLemma1}. 
\end{proof}

Note that the uniform approximation assumption in (\ref{UniformAssumption}) is satisfied by losses defined in terms of geodesic distance and is therefore not merely an abstract technical restriction. To elaborate more on this, we give the following examples.

\textit{Example 1 (Squared-geodesic loss)}:
Consider the intrinsic squared-geodesic loss
\begin{eqnarray}
	R_i(\bm w)= R_i(\bm\Theta_i)+\frac{\lambda_i}{2} d_{\mathrm{geo}}(\bm w,\bm\Theta_i)^2, \qquad
\lambda_i>0.
\end{eqnarray}
This loss has its unique global minimum at $\bm\Theta_i$. In the local geodesic coordinates, mapping $E_{\bm{\Theta}_i}$, introduced in Appendix \ref{ProofLemma1}, write $\bm{w} = E_{\bm{\Theta}_i}(\bm{v})$ for $\bm{v}\in \mathcal{T}_{\bm{\Theta}_i}$. By the defining property of geodesic coordinates, $d_{\mathrm{geo}}(\bm w,\bm\Theta_i)=\|\bm v\|$, and hence the loss expressed in local geodesic coordinates is

\begin{eqnarray}
	f_i(\bm v)=f_i(\bm 0)+\frac{\lambda_i}{2}\|\bm v\|^2.
\end{eqnarray}
Consequently, with $\bm v=\sqrt d\,\bm z$, it yields
\begin{eqnarray}
	\frac{1}{d}\left(f_i(\sqrt d\,\bm z)-f_i(\bm 0)\right)= \frac{\lambda_i}{2}\|\bm z\|^2.
\end{eqnarray}
Thus, the uniform quadratic approximation assumption holds exactly
\begin{eqnarray}
	\sup_{\|\bm z\|\leq C}\left|\frac{1}{d}\left(f_i(\sqrt d\,\bm z)-f_i(\bm 0)\right)-\frac{\lambda_i}{2}\|\bm z\|^2 \right| =0
\end{eqnarray}
for every fixed $C<\pi$. In particular, $\nabla^2 f_i(\bm 0)=\lambda_i\mathbf I_{d-1}$.

\textit{Example 2 (Nonlinear perturbations)}:
The assumption also permits nonlinear losses, provided that the nonlinear contribution is of lower order than $d$ on the relevant $\|\bm v\|=O(\sqrt d)$ scale. For example, consider the family
\begin{eqnarray}
	R_i^{(d)}(\bm w)=R_i^{(d)}(\bm\Theta_i)+\frac{\lambda_i}{2}d_{\mathrm{geo}}(\bm w,\bm\Theta_i)^2+\frac{a_i}{d^{1+\delta}}d_{\mathrm{geo}}(\bm w,\bm\Theta_i)^4,\qquad\delta>0,
\end{eqnarray}
where $a_i\in\mathbb R$ is fixed. In the same local geodesic coordinates, we have
\begin{eqnarray}
	f_i(\bm v)=f_i(\bm 0)+\frac{\lambda_i}{2}\|\bm v\|^2 + \frac{a_i}{d^{1+\delta}}\|\bm v\|^4 .
\end{eqnarray}
Again by setting $\bm v=\sqrt d\bm z$, we get
\begin{eqnarray}
	\frac{1}{d}\left(f_i(\sqrt d\,\bm z)-f_i(\bm 0)\right)=\frac{\lambda_i}{2}\|\bm z\|^2 +\frac{a_i}{d^\delta}\|\bm z\|^4,
\end{eqnarray}
resulting into 
\begin{eqnarray}
	\sup_{\|\bm z\|\leq C} \left|\frac{1}{d}\left(f_i(\sqrt d\,\bm z)-f_i(\bm 0)\right)-
\frac{\lambda_i}{2}\|\bm z\|^ 2 \right|\leq
\frac{|a_i|C^4}{d^\delta}
\longrightarrow 0,
\end{eqnarray}
as $d\rightarrow \infty$ for every fixed $C<\pi$. Hence, the uniform quadratic approximation assumption remains valid despite the presence of a non-quadratic term. Moreover, the degree four perturbation does not change the Hessian at the minimizer, so that $\nabla^2 f_i(\bm 0)=\lambda_i\mathbf I_{d-1}$.

\begin{remark}
	It is shown in \cite{Lanzillotta2023} that a higher rank of the average Hessian matrices over the first $t-1$ tasks corresponds to a lower capability of learning the subsequent task, meaning task $t$. Here, the assumption that the Hessian matrices $\mathbf{H}_i$ are isotropic ensures that their average over the first $t-1$ tasks remains full rank. Therefore, the following results in this paper are under the assumption of the worst-case scenario for the capability of learning a new task. 
\end{remark}

\section{Task Optima as Neyman-Scott Clustered Point Process} \label{sec2}
In continual learning, tasks arrive sequentially and are often heterogeneous but not necessarily independent of one another. In particular, different tasks may share latent structure, leading to groups of tasks that are more closely related than tasks belonging to other groups. This observation has motivated a number of approaches that explicitly model task heterogeneity through task similarity and clustering. For example, \cite{Yao2019} introduced a hierarchical task-clustering framework to address task uncertainty and heterogeneity, with the goal of customizing transferable knowledge to different clusters while encouraging knowledge sharing among tasks within the same cluster. Their extension to continual learning further recognizes that task relationships may evolve as new tasks arrive \cite{Yao2019}. Moreover, \cite{Peng2026} argued that understanding tasks dependencies on one another is important for a theoretical view on continual learning. 

We borrow the same assumption that tasks are not independent; they form clusters. Within each cluster, tasks are more similar to one another than the tasks from other clusters, i.e., tasks within a cluster are dependent statistically to their cluster center while the task optima belonging to different clusters are independent.
To mathematically formulate the grouping of correlate tasks in this paper, we model the distribution of task optima $\bm{\Theta}_i$ as a spherical Neyman-Scott cluster point process via a two-tired generative hierarchy:
\subsubsection{Parent Process}
A set of parent points $\{\bm{Y}_{j}\}_{i=1}^{N_p}$ is generated via a homogeneous Poisson point process $\Phi_p$ on $\mathbb{S}^{d-1}(\sqrt{d})$ with intensity $\lambda$ such that for every measurable $\mathcal{A}\subseteq \mathbb{S}^{d-1}(\sqrt{d})$, the number of parent points in $\mathcal{A}$, denoted by $\Phi_p(\mathcal{A})$, are distributed by $\mathrm{Poisson}\left(\lambda \sigma(\mathcal{A}) \left| \mathbb{S}^{d-1}(\sqrt{d}) \right| \right)$, where $\left| \mathbb{S}^{d-1}(\sqrt{d}) \right|$ is the surface area of hypersphere. For disjoint measurable sets $\mathcal{A}_{1}, ..., \mathcal{A}_k$, random variables $\Phi_p(\mathcal{A}_1), ..., \Phi_p(\mathcal{A}_k)$ are independent. Each parent point $\bm{Y}_{j}$ represents its corresponding cluster center. Note that the parent points are not part of the Neyman–Scott process; they are simply the invisible centers of the clusters. 

\subsubsection{Offspring Process}	
Given the parent process $\Phi_p$, for each given parent point $\bm{Y}_j$, there is a cluster of size $C_j\sim\mathrm{Poisson}(c)$ of offspring points $\bm{\Theta}_i$, sampled independently from a von Mises-Fisher (vMF) distribution with mean direction $\bm{Y}_j$ and concentration parameter $\kappa$, denoted by $\mathrm{vMF}(\bm{Y}_j, \kappa)$. Specifically, conditional on $\bm{Y}_j$, the density of an offspring point $\bm{\Theta}_i\in\mathbb{S}^{d-1}(\sqrt d)$ with respect to Haar measure on $\mathbb{S}^{d-1}(\sqrt d)$ is
\begin{eqnarray}
		f_{\bm{\Theta}_i|\bm{Y}_j}\left(\bm{\theta}_{i}\middle \vert \bm{y}_j\right) = \frac{C_d(\kappa)}{d^{(d-1)/2}}\exp\left\{\frac{\kappa}{d} \langle \bm{y}_j,\bm{\theta}_i \rangle \right\}, \label{vMF}
	\end{eqnarray}
	where 
	\begin{eqnarray}
		C_d(\kappa)= \frac{\kappa^{d/2-1}}{(2\pi)^{d/2}I_{d/2-1}(\kappa)},\label{Cd} 
	\end{eqnarray}
	with $I_{d/2-1}(.)$ being the modified Bessel function of first kind with order $d/2-1$. Note that every offspring point sampled from (\ref{vMF}) also lies on $\mathbb{S}^{d-1}(\sqrt{d})$. The vMF distribution may be viewed as the spherical counterpart of the Gaussian distribution \cite{Mardia1999} used for offspring generation in Euclidean Neyman-Scott processes.
	
	We denote the collection of sampled offspring points from vMF distribution with mean directions  $\{ \bm{Y}_{j}\}_{j=1}^{N_p}$ as the Neyman-Scott point process
	\begin{eqnarray}
		\Phi = \left\{ \bm{\Theta}_1,..., \bm{\Theta}_i,..., \bm{\Theta}_{|\Phi|}\right\},
	\end{eqnarray}
	where $|\Phi| = \sum_{j=1}^{N_p}C_j$. Note that here $N_p \sim \mathrm{Poisson}\left(\lambda \left| \mathbb{S}^{d-1}(\sqrt{d})\right|\right)$ and conditionally on $N_p$, the number of points in each cluster $C_j \sim \mathrm{Poisson}(c)$. 
	\begin{proposition}(Typical Geodesic Distance   of an Offspring from its Parent)\label{Prop1}
		Given $N_p$, let $\bm{\Theta}_i$ be a point sampled from the Neyman-Scott point process $\Phi$ with parent $\bm{Y}_{j}$. As $\kappa , d \rightarrow \infty$ such that $\kappa / d \rightarrow \alpha$, where $\alpha >0$ is a constant, the ratio $d_{geo}(\bm{\Theta}_i,\bm{Y}_j)/\sqrt{d}$ converges in probability to a constant 
		\begin{eqnarray}
			\frac{d_{geo}(\bm{\Theta}_i,\bm{Y}_j)}{\sqrt{d}} \xrightarrow{P} \arccos u_{\alpha}^\prime , 
		\end{eqnarray} 
		where 
		\begin{eqnarray}
			u_{\alpha}^\prime =	\frac{\sqrt{1+4\alpha^2}-1}{2\alpha}.
		\end{eqnarray}
	\end{proposition}
\begin{proof}
Proof is given in Appendix \ref{ProofProp1}.
\end{proof}

\begin{remark}\textit{(Control of Task Optima Clusters Using $\alpha$)}
	Given $N_p$, the points generated by a homogeneous Poisson point process become uniformly distributed on $\mathbb{S}^{d-1}(\sqrt{d})$. It is known that, for any pair of points uniformly distributed on a high dimensional sphere, their normalized inner product converges to zero in probability as $d \to \infty$, meaning that the points become asymptotically orthogonal \cite{Cai2013}. In the present setting, Proposition \ref{Prop1} reveals two distinct regimes governed by the relative growth of the dimension and the concentration parameter.
When the dimension grows faster than the concentration parameter, corresponding to $\alpha \to 0$, we have
\begin{eqnarray}
	\frac{d_{geo}(\bm{\Theta}_i,\bm{Y}_j)}{\sqrt{d}} \xrightarrow{P} \frac{\pi}{2}.
\end{eqnarray}
Thus, offspring points become asymptotically orthogonal to their parents and effectively lose their dependence on them. Instead, they behave as points drawn from the uniform distribution of the parent population. Consequently, the offspring no longer exhibit a localized cluster structure.
In contrast, when the concentration parameter grows faster than the dimension, corresponding to $\alpha \to \infty$, the ratio
\begin{eqnarray}
	\frac{d_{geo}(\bm{\Theta}_i,\bm{Y}_j)}{\sqrt{d}} \xrightarrow{P} 0.
\end{eqnarray}
In this regime, offspring points become tightly concentrated around their respective parents, preserving a strong parent-offspring dependence and producing well-defined clusters.
\end{remark}
	
\section{Continual Learning Capacity under Non-Forgetting Constraint}\label{sec3}

The ultimate goal of continual learning is for a learner to acquire a sequence of $T$ tasks without forgetting the previous $T-1$ tasks upon arriving the $T$-th task. Here, the sequence of tasks is modeled by drawing uniformly and without replacement a set of $T$ points (task optima) from $\Phi$ on $\mathbb{S}^{d-1}(\sqrt{d})$. 

\begin{remark}(\textit{On Cardinality of $ \Phi $})
	 As mentioned before, $|\Phi|=\sum_{j=1}^{N_p}C_j$, where $N_p\sim\mathrm{Poisson}\left(\lambda\left|\mathbb S^{d-1}(\sqrt d)\right|\right)$ and, conditionally on $N_p$, the number of points in each cluster satisfies
$C_j\sim\mathrm{Poisson}(c)$. Thus, $ |\Phi|$ is a compound Poisson random variable with expectation
\begin{eqnarray}
	\mathbb E[|\Phi|]=\mathbb E[C_j]\mathbb E[N_p]=c\lambda\left|\mathbb S^{d-1}(\sqrt d)\right|,
\end{eqnarray}
and variance
\begin{eqnarray}
	\operatorname{Var}(|\Phi|)=\mathbb E[N_p]\operatorname{Var}(C_j)
+\operatorname{Var}(N_p)\mathbb E[C_j]^2=c\lambda\left|\mathbb S^{d-1}(\sqrt d)\right|(1+c).
\end{eqnarray}

Now, $T$ is fixed and does not depend on $d$. Since $\left|\mathbb S^{d-1}(\sqrt d)\right|\rightarrow\infty$ as $d\to\infty$, we have $c\lambda\left|\mathbb S^{d-1}(\sqrt d)\right|\rightarrow\infty$.
Consequently, for any fixed $\epsilon>0$, there exists $d_0$ such that, for all $d\ge d_0$,
\begin{eqnarray}
	T\le (1-\epsilon)c\lambda\left|\mathbb S^{d-1}(\sqrt d)\right|.
\end{eqnarray}
Hence, for all sufficiently large \(d\), Chebyshev's inequality gives
\begin{eqnarray}
	\mathbb P[|\Phi|<T]\!\leq\! \mathbb P\left[|\Phi|<(1-\epsilon)c\lambda\left|\mathbb S^{d-1}(\sqrt d)\right|\right]\!\leq\! \frac{\operatorname{Var}(|\Phi|)}{\epsilon^2c^2\lambda^2 \left|\mathbb S^{d-1}(\sqrt d)\right|^2}=\frac{1}{\epsilon^2}
\left(\frac{1}{\lambda\left|\mathbb S^{d-1}(\sqrt d)\right|}+\frac{1}{c\lambda\left|\mathbb S^{d-1}(\sqrt d)\right|}\right)\rightarrow 0.
\end{eqnarray}
Thus, since $T$ is fixed while
$c\lambda|\mathbb S^{d-1}(\sqrt d)|\to\infty$, the required assumption is satisfied for all sufficiently large $d$, and $\mathbb P[|\Phi|<T]\longrightarrow 0$, i.e., we can choose any $T$-subset of $\Phi$ for any $T$ as long as $T$ is a constant.
\end{remark}

\subsection{Sequence of Channels and Number of Tasks without Forgetting}
Let the intersection of the acceptable loss regions for the first $t$ tasks be

\begin{eqnarray}
	\mathcal{K}_{t}(\bm{\Theta}_{1:t}) := \bigcap_{i=1}^{t} B_{\bm{\Theta}_{i}}(r_i(d)),
\end{eqnarray}
with $\mathcal{K}_0 = \mathbb{S}^{d-1}(\sqrt{d})$. Throughout this paper we refer to $\mathcal{K}_{t}(\bm{\Theta}_{1:t})$ as the \textit{feasible set} of the first $t$ tasks.

The learning process of task $t\in \{1, ..., T\}$ is considered as a channel (Markov kernel) 
\begin{eqnarray}
	Q_t(\mathrm{d}\bm{w}_t|\bm{w}_{t-1},\mathcal{K}_{t-1}(\bm{\theta}_{1:t-1}), \bm{\theta}_t),
\end{eqnarray}
where $\bm{W}_{t-1}=\bm{w}_{t-1}$ denotes the parameter learned from the previous task (task $t-1$), with the initial parameter $\bm{W}_0$ uniformly distributed on the sphere, i.e., $\bm{W}_0\sim\sigma$, and $\bm{\Theta}_t=\bm{\theta}_t$ denotes the optimum parameter for the current task. The channel $Q_t$ corresponding to task $t$ outputs the parameter $\bm{W}_t$ for task $t$ (i.e., the next task) such that
\begin{eqnarray}
	\bm{W}_t \sim Q_t\left(. \middle \vert \bm{W}_{t-1}, \mathcal{K}_{t-1}(\bm{\Theta}_{1:t-1}), \bm{\Theta}_t\right). 
\end{eqnarray}
Note that channels $Q_t$ are probability measures on sphere. 
Each $Q_t$ is absolutely continuous with respect to $\sigma$, meaning there exists the density $q_t(\bm{w}_t|\bm{w}_{t-1}, \mathcal{K}_{t-1}(\bm{\theta}_{1:t-1}), \bm{\theta}_t)$ such that
\begin{eqnarray}
	q_t(\bm{w}_t|\bm{w}_{t-1}, \mathcal{K}_{t-1}(\bm{\theta}_{1:t-1}), \bm{\theta}_t)=\frac{Q_t(\mathrm{d}\bm{w}_t|\bm{w}_{t-1}, \mathcal{K}_{t-1}(\bm{\theta}_{1:t-1}), \bm{\theta}_t)}{\sigma(\mathrm{d}\bm{w}_t)}. 
\end{eqnarray} 
For each $t \in \{1, ..., T\}$ and the corresponding constant $\rho_t>0$, we impose 
\begin{eqnarray}
	D(Q_t(.| \bm{w}_{t-1},\mathcal{K}_{t-1}(\bm{\theta}_{1:t-1}), \bm{\theta}_t) \| \sigma) \leq \rho_t,\label{KL}
\end{eqnarray}
where 
\begin{eqnarray}
	D(Q_t \| \sigma) = \int_{\mathbb{S}^{d-1}(\sqrt{d})}q_t(\bm{w}_t|\bm{w}_{t-1},\mathcal{K}_{t-1},\bm{\theta}_{t}) \log q_t(\bm{w}_t|\bm{w}_{t-1},\mathcal{K}_{t-1},\bm{\theta}_{t}) \sigma(\mathrm{d}\bm{w}_t)
\end{eqnarray}
is the KL divergence. The KL constraint allows compactly supported densities and forbids arbitrarily concentrated deterministic channels.

To avoid forgetting upon the arrival of a new task and to ensure that the new task is learned with an acceptable error, we require that, for every $t$, the learned parameter of the new task, $\bm{W}_t$, lies in the feasible set of $t-1$ previous tasks as well as the acceptable loss region of current task $B_{\bm{\Theta}_t}(r_t(d))$ with high probability, i.e.,
\begin{eqnarray}
	\mathbb{P}\left[\bm{W}_t\in B_{\bm{\Theta}_t}(r_t(d))\cap \mathcal{K}_{t-1}(\bm{\Theta}_{1:t-1})\right]=\mathbb{P}\left[\bm{W}_t \in \mathcal{K}_{t}(\bm{\Theta}_{1:t})\right] \geq 1-\delta_t, \qquad \forall t\in \{1,..., T\},\label{feasibility}
\end{eqnarray}
for any $\delta_t>0$. However, for $\bm{W}_t$ to satisfy (\ref{feasibility}), we must ensure that $\mathcal{K}_{t}(\bm{\Theta}_{1:t})$ is non-empty for all $t=1,\ldots,T$ with high probability as $d\rightarrow\infty$. Hence, before proceeding with the sequence of channels, we at first derive an upper bound on $T$ such that, for any $T$ below this bound, the feasible region is non-empty with high probability as $d\rightarrow\infty$.

\begin{theorem}\label{Thm1}
	Let $\{ \bm{\Theta}_1, ..., \bm{\Theta}_T\}$ be $T$ task optima chosen uniformly without replacement from $\Phi$. The loss region corresponding to each task optima is a geodesic ball $B_{\bm{\Theta}_i}(r_i(d))$ centered at $\bm{\Theta}_i$ with radius $r_i(d)/\sqrt{d}= \sqrt{2\eta/\lambda_i}$ as in Lemma \ref{Lemma1}. Let $\lambda^* = \max_{1\leq i\leq T} \lambda_i$ and assume $\min_{1\leq i\leq T} r_i(d)/\sqrt{d} > \arccos u_\alpha^\prime$. If 
	\begin{eqnarray}
		T< \frac{1}{\cos^2\left(\sqrt{\frac{2\eta}{\lambda^*}} -\arccos u^\prime_\alpha\right)},\label{Tbound}
	\end{eqnarray}
	then, $\mathbb{P}\left[\mathcal{K}_{T}(\bm{\Theta}_{1:T}) \neq \emptyset\right]\rightarrow 1$ as $d\rightarrow \infty$ and $\kappa/d \rightarrow \alpha$.
\end{theorem}
\begin{proof}
	Proof is given in Appendix \ref{ProofThm1}. 
\end{proof}

Let $T$ satisfies (\ref{Tbound}). For a fixed $t\in \{1, ..., T\}$, given previous task learned parameter $\bm{W}_{t-1}=\bm{w}_{t-1}$, the feasible set of all previous tasks $\mathcal{K}_{t-1}(\bm{\Theta}_{1:t-1})=\mathcal{K}_{t-1}(\bm{\theta}_{1:t-1})$, and current optimum $\bm{\Theta}_t=\bm{\theta}_t$, the probability that task $t$ is learned and forgetting is not happened is $Q_t(\mathcal{K}_{t}(\bm{\theta}_{1:t}) | \bm{w}_{t-1}, \mathcal{K}_{t-1}, \bm{\theta}_t)$. We then pose the question: under the KL divergence restriction in (\ref{KL}), what is the largest possible probability 
\begin{eqnarray}
	Q_t\left(\mathcal{K}_{t}(\bm{\theta}_{1:t}) \middle \vert \bm{w}_{t-1}, \mathcal{K}_{t-1}, \bm{\theta}_t\right) = \int_{\mathcal{K}_{t}(\bm{\theta}_{1:t})}q_t(\bm{w}_t | \bm{w}_{t-1},\mathcal{K}_{t-1},\bm{\theta}_t)\sigma(\mathrm{d}\bm{w}_t)
\end{eqnarray}
that the channel can place on $\mathcal{K}_{t}(\bm{\theta}_{1:t})$? To answer this, fix $d$ and let $\nu_t := \sigma \left(\mathcal{K}_{t}(\bm{\theta}_{1:t})\right)$ be the normalized Haar measure of feasible set $\mathcal{K}_t$. we denote
\begin{eqnarray}
	\psi_{\rho_t}(\nu_t)&& \!\!\!\!\!\!\!\!\!\!\!:= \sup_{D(Q_t \| \sigma) \leq \rho_t} Q_t(\mathcal{K}_{t}(\bm{\theta}_{1:t})| \bm{w}_{t-1},\mathcal{K}_{t-1},\bm{\theta}_t) \nonumber \\
	&&\!\!\!\!\!\!\!\!\!\!\!  = \sup_{q_t(.|\bm{w}_{t-1},\mathcal{K}_{t-1},\bm{\theta}_t): D(Q_t \| \sigma) \leq \rho_t} \ \ \int_{\mathcal{K}_{t}(\bm{\theta}_{1:t})}q_t(\bm{w}_t|\bm{w}_{t-1},\mathcal{K}_{t-1},\bm{\theta}_t)\sigma(\mathrm{d}\bm{w}_t). \label{psi}
\end{eqnarray}
Let $q_t$ be any density satisfying the KL restriction in (\ref{KL}). Define,
\begin{eqnarray}
	x_t := \int_{\mathcal{K}_{t}(\bm{\theta}_{1:t})}q_t(\bm{w}_t|\bm{w}_{t-1},\mathcal{K}_{t-1},\bm{\theta}_t)\sigma(\mathrm{d}\bm{w}_t).\label{x} 
\end{eqnarray}
Then, it is clear that
\begin{eqnarray}
	1-x_t := \int_{\mathcal{K}_{t}(\bm{\theta}_{1:t})^{c}}q_t(\bm{w}_t|\bm{w}_{t-1},\mathcal{K}_{t-1},\bm{\theta}_t)\sigma(\mathrm{d}\bm{w}_t).
\end{eqnarray}
Now, we define a new density $\bar{q}_t$ that is constant on both $\mathcal{K}_{t}(\bm{\theta}_{1:t})$ and its complement $\mathcal{K}_{t}(\bm{\theta}_{1:t})^c$ as
\begin{eqnarray}
	\bar{q}_{t}(\bm{w}_t|\bm{w}_{t-1},\mathcal{K}_{t-1},\bm{\theta}_t) = 
	\begin{cases}
		\frac{x_t}{\nu_t}, & \bm{w}_t\in \mathcal{K}_{t}(\bm{\theta}_{1:t})\\
		\frac{1-x_t}{1-\nu_t}, & \bm{w}_t \notin \mathcal{K}_{t}(\bm{\theta}_{1:t})
	\end{cases}.\label{barq}
\end{eqnarray}
By construction, we have
\begin{eqnarray}
	\int_{\mathcal{K}_{t}(\bm{\theta}_{1:t})}\bar{q}_t(\bm{w}_t|\bm{w}_{t-1},\mathcal{K}_{t-1},\bm{\theta}_t) \sigma(\mathrm{d}\bm{w}_t) = x_t. \label{barqxt}
\end{eqnarray}
So, $\bar{q}_t$ gives the same mass on $\mathcal{K}_t(\bm{\theta}_{1:t})$ as the original density $q_t$. Since function $g(s)=s \log s$ is convex for $s>0$, then by Jensen's inequality on $\mathcal{K}_t(\bm{\theta}_{1:t})$, we get
\begin{eqnarray}
	\nu_t g\left(\frac{x_t}{\nu_t}\right)\leq \int_{\mathcal{K}_{t}(\bm{\theta}_{1:t})}q_t(\bm{w}_t|\bm{w}_{t-1},\mathcal{K}_{t-1},\bm{\theta}_t) \log q_t(\bm{w}_t|\bm{w}_{t-1},\mathcal{K}_{t-1},\bm{\theta}_t) \sigma(\mathrm{d}\bm{w}_t).\label{40}
\end{eqnarray}
Similarly on $\mathcal{K}_{t}(\bm{\theta}_{1:t})^c$, by the same procedure, we have
\begin{eqnarray}
	(1-\nu_t)g\left(\frac{1-x_t}{1-\nu_t}\right)\leq \int_{\mathcal{K}_{t}(\bm{\theta}_{1:t})^c}q_t(\bm{w}_t|\bm{w}_{t-1},\mathcal{K}_{t-1},\bm{\theta}_t) \log q_t(\bm{w}_t|\bm{w}_{t-1},\mathcal{K}_{t-1},\bm{\theta}_t) \sigma(\mathrm{d}\bm{w}_t).\label{41}
\end{eqnarray}
From (\ref{40}) and (\ref{41}) we can write
\begin{eqnarray}
	D(Q_t \| \sigma) = \int_{\mathcal{K}_t}q_t \log q_t \mathrm{d}\sigma + \int_{\mathcal{K}_t^c} q_t\log q_t \mathrm{d}\sigma \geq x_t\log \frac{x_t}{\nu_t}+(1-x_t)\log \frac{1-x_t}{1-\nu_t}= D(\bar{Q}_t \| \sigma).\label{42}
\end{eqnarray}
We have shown that for any $q_t$ satisfying the KL condition, there exists a two-level density $\bar{q}_t$ with the same mass on $\mathcal{K}_t$ that also satisfies the KL condition 
\begin{eqnarray}
	D(\bar{Q}_t \| \sigma) \leq D(Q_t \| \sigma) \leq \rho_t,
\end{eqnarray}
as in (\ref{42}). Thus, every achievable mass $x_t$ is achievable by a two-level density. Therefore, we lose nothing by restricting the optimization in (\ref{psi}) to two-level densities. The set of achievable masses $x_t$ is then $\left\{x_t\in[\nu_t,1): D(\bar{Q}_t \| \sigma ) \leq \rho_t \right\}$. We know that $D(\bar{Q}_t \| \sigma)$ (a function of $x_t$ as in (\ref{42})) is continuous and increasing on $[\nu_t,1)$ with $D(\bar{Q}_t \| \sigma) =0$ when $x_t=\nu_t$ and $D(\bar{Q}_t \| \sigma) \rightarrow \infty$ when $x_t \rightarrow 1$. Hence, the maximum achievable mass is
\begin{eqnarray}
	\psi_{\rho_t}(\nu_t) = x_t^*,
\end{eqnarray}
where $x_t^*$ satisfying 
\begin{eqnarray}
	x_t^* \log \frac{x_t^*}{\nu_t}+(1-x_t^*)\log \frac{1-x_t^*}{1-\nu_t}=\rho_t.
\end{eqnarray}
Therefore, we obtain
\begin{eqnarray}
	Q_t(\mathcal{K}_{t}(\bm{\theta}_{1:t})|\bm{w}_{t-1},\mathcal{K}_{t-1},\bm{\theta}_t) \leq \psi_{\rho_t}(\sigma(\mathcal{K}_{t}(\bm{\theta}_{1:t}))). \label{UPPonQ_t}
\end{eqnarray}
For task $t$, the non-forgetting and reliable probability as mentioned in (\ref{feasibility}) is
\begin{eqnarray}
	\mathbb{P}\left[\bm{W}_t\in \mathcal{K}_{t}(\bm{\Theta}_{1:t})\right] = \mathbb{E}_{\bm{\Theta}_{1:t}, \bm{W}_{t-1}}\left[Q_t(\mathcal{K}_{t}(\bm{\Theta}_{1:t})|\bm{W}_{t-1},\mathcal{K}_{t-1},\bm{\Theta}_t)\right].
\end{eqnarray}
Since (\ref{UPPonQ_t}) holds for every $\bm{w}_{t-1}$, it is also holds after averaging over $\bm{W}_{t-1}$ and over the task optima $\{ \bm{\Theta}_1, ..., \bm{\Theta}_t\}$; hence, we have 
\begin{eqnarray}
	\mathbb{P}\left[\bm{W}_t\in \mathcal{K}_{t}(\bm{\Theta}_{1:t})\right] \leq \mathbb{E}_{\bm{\Theta}_{1:t}}\left[\psi_{\rho_t}(\sigma(\mathcal{K}_{t}(\bm{\Theta}_{1:t})))\right],
\end{eqnarray}
which results into
\begin{eqnarray}
	\mathbb{E}_{\bm{\Theta}_{1:t}}\left[\psi_{\rho_t}(\sigma(\mathcal{K}_{t}(\bm{\Theta}_{1:t})))\right] \geq 1-\delta_t, \qquad \forall t\in\{1, ..., T\}.\label{rhotdelta}
\end{eqnarray}
Building upon the result in (\ref{rhotdelta}), we further characterize the capacity for continual learning.

\subsection{Continual Learning Capacity}
We define the admissible learning process class (class of channels) for tasks $t\in\{1,..., T\}$  as
\begin{eqnarray}
	\mathcal{Q}\left(\rho_{1:T}, \delta_{1:T}\right):=\left\{Q_{1:T} :
\begin{aligned}
&D\left(Q_t(\cdot \mid \bm{w}_{t-1}, \mathcal{K}_{t-1}, \bm{\theta}_t)\| \sigma\right) \leq \rho_t,
&& t=1,\ldots,T,\\
&\mathbb{P}\left[\bm{W}_t \in \mathcal{K}_t(\bm{\Theta}_{1:t})\right] \geq 1-\delta_t,
&& t=1,\ldots,T
\end{aligned}
\right\}.
\end{eqnarray}
The \textit{capacity of continual learning} for a fixed $d$ and the sequence of $T$ tasks is defined as
\begin{eqnarray}
	C_T(\rho_{1:T},\delta_{1:T},d) := \sup_{Q_{1:T}\in \mathcal{Q}(\rho_{1:T}, \delta_{1:T})} I\left(\bm{\Theta}_{1:T}; \bm{W}_{1:T}\right), \label{capacity}
\end{eqnarray}
where $I(.;.)$ is the mutual information. Note that here, we consider causal channels, meaning that $\bm{W}_t$ is conditionally independent of $\bm{\Theta}_{t+1:T}$ given $\bm{\Theta}_{1:t}$ and $\bm{W}_{1:t-1}$. 
\begin{theorem}\label{Thm2}
	Let $u_t = \sigma\left(\bigcup_{t^\prime=1}^{t}\mathcal{K}_{t^\prime}(\bm{\theta}_{1:t^\prime})\right)$. The continual learning capacity define in (\ref{capacity}) is upper bounded by
	\begin{eqnarray}
		C_T(\rho_{1:T}, \delta_{1:T},d) \leq \sum_{t=1}^{T}\left(\inf_{\rho_t \in \mathcal{S}_{\rho_t,\delta_t}}\rho_t - d_{+}(1-\delta_t \| u_t)\right),
	\end{eqnarray}
	where $\mathcal{S}_{\rho_t,\delta_t}$ is the set of all $\rho_t$ satisfying (\ref{rhotdelta}), i.e., for all $t=1, ..., T$,
	\begin{eqnarray}
		\mathcal{S}_{\rho_t,\delta_t}=\left\{\rho_t :\mathbb{E}_{\bm{\Theta}_{1:t}}\left[\psi_{\rho_t}(\sigma(\mathcal{K}_{t}(\bm{\Theta}_{1:t})))\right]\geq 1-\delta_t\right\},
	\end{eqnarray}
	and 
	\begin{eqnarray}
		d_{+}(1-\delta_t \| u_t) = \begin{cases}
		D_{\mathrm{bin}}(1-\delta_t \| u_t), & u_t < 1-\delta_t\\
		0, & u_t\geq 1-\delta_t
	\end{cases}.
	\end{eqnarray}
	Here, $D_{\mathrm{bin}}(p\| q) = p\log \frac{p}{q}+(1-p)\log \frac{1-p}{1-q}$ is the binary relative entropy. If we set $\delta_t=0, \ \forall t=1,...,T$, then
	\begin{eqnarray}
		C_T(\rho_{1:T},0,d) \leq \sum_{t=1}^{T}\left(\inf_{\rho_t\in\mathcal{S}_{\rho_t,0}}\rho_t -\log \frac{1}{u_t}\right).
	\end{eqnarray}
\end{theorem}
\begin{proof}
	Proof is given in Appendix \ref{ProofThm2}.
\end{proof}

\textit{Example 3 (Capacity of Simple Channels)}:  Consider simple channels (learning process) for tasks $t\in\{1,...,T\}$ that outputs a point uniformly distributed over $\mathcal{K}_{t}(\bm{\theta}_{1:t})$. Formally,
\begin{eqnarray}
	Q_t(\mathrm{d}\bm{w}_t|\bm{w}_{t-1}, \mathcal{K}_{t-1},\bm{\theta}_t) = \frac{\mathbf{1}_{\mathcal{K}_t(\bm{\theta}_{1:t})}(\bm{w}_t)}{\sigma(\mathcal{K}_t(\bm{\theta}_{1:t}))}\sigma(\mathrm{d}\bm{w}_t), \qquad \forall t=1,...,T.\label{simplechannel}
\end{eqnarray}
We have already shown that if any channel $Q_t$ satisfying $Q_t(\mathcal{K}_{t}(\bm{\theta}_{1:t})|\bm{w}_{t-1},\mathcal{K}_{t-1},\bm{\theta}_t)=x_t$ as in (\ref{x}) and $D(Q_t\| \sigma) \leq \rho_t$, then replacing $q_t$ by a two-level density $\bar{q}_t$ as in (\ref{barq}) does not increase the KL divergence and therefore, $D_{\mathrm{bin}}(x_t\| \nu_t) \leq \rho_t$. Conversely, suppose $D_{\mathrm{bin}}(x_t\| \nu_t) \leq \rho_t$. Then, we can construct the two-level density $\bar{q}_t$ with $\bar{Q}_t(\mathcal{K}_{t}(\bm{\theta}_{1:t})|\bm{w}_{t-1},\mathcal{K}_{t-1},\bm{\theta}_t)= x_t$ and $D(\bar{Q}_t \| \sigma)=D_{\mathrm{bin}}(x_t\| \nu_t) \leq \rho_t$. Thus, $\bar{Q}_t$ is admissible and achieves mass $x_t$ on $\mathcal{K}_t(\bm{\theta}_{1:t})$. Then, we can conclude that $x_t$ is achievable under budget $\rho_t$ if and only if $D_{\mathrm{bin}}(x_t\| \sigma) \leq \rho_t$. So, the largest achievable mass is 
\begin{eqnarray}
	\psi_{\rho_t}(\nu_t) = \max\left\{x_t\in [\nu_t,1]: D_{\mathrm{bin}}(x_t\| \nu_t) \leq \rho_t\right\}. 
\end{eqnarray}
It is simple to observe that: $D_{\mathrm{bin}}(x_t\| \nu_t)$ is increasing in $x_t$ for $x_t>\nu_t$; at $x_t=\nu_t$, we have $D_{\mathrm{bin}}(x_t\| \nu_t)=0$; and as $x_t\rightarrow 1$, we have $D_{\mathrm{bin}}(x_t\| \nu_t) \rightarrow \log \frac{1}{\nu_t}$. Hence, if $\rho_t < \log \frac{1}{\nu_t}$, then $\psi_{\rho_t}(\nu_t)$ is the unique $x_t^*\in (\nu_t,1)$ satisfying $D_{\mathrm{bin}}(x_t^*\| \nu_t)=\rho_t$. If $\rho_t \geq \log \frac{1}{\nu_t}$, then $x_t=1$ is feasible and therefore $\psi_{\rho_t}(\nu_t)=1$, i.e.,
\begin{eqnarray}
	\psi_{\rho_t}(\nu_t) = \begin{cases}
		x_t^*, & 0<\rho_t<\log \frac{1}{\nu_t}\\
		1, & \rho_t \geq \log \frac{1}{\nu_t}
	\end{cases}.
\end{eqnarray}
Thus, when $\rho_t < \log \frac{1}{\nu_t}$, it is not possible to place all mass on $\mathcal{K}_{t}(\bm{\theta}_{1:t})$. 

We observe that the budget control $\rho_t$ depends on $\nu_t = \sigma \left(\mathcal{K}_{t}(\bm{\theta}_{1:t})\right)$. This naturally raises the question: what can be said about $\nu_t$? 
To answer this, let's go back to high dimensional clustering regime $\kappa / d \rightarrow \alpha$ as $\kappa , d\rightarrow \infty$ and the underlying geometry. 

The number of offspring points (offspring count) in each cluster $C_j$ are independent of the parent locations. Thus, the offspring counts can affect \emph{which parent contributes to more offspring points}, but they do not affect the spatial distribution of parent locations, i.e., conditional on the offspring counts, the parent locations remain iid Haar
\begin{eqnarray}
	(\bm{Y}_1,\ldots,\bm{Y}_{N_p})\mid (C_1,\ldots,C_{N_p})
\stackrel{\mathrm{i.i.d}}{\sim}\sigma.
\end{eqnarray}
Consequently, if we select $t$ offspring points with \emph{distinct} parent labels $J_1,...,J_{M_t}$ where $1\leq M_t \leq t$, then the corresponding distinct parent locations is still uniformly distributed on $\mathbb{S}^{d-1}(\sqrt{d})$, although parent $J_j$ with larger offspring count $C_j$ might represent more offspring points among the chosen $t$ points. Note that here the number of distinct parent points $1\leq M_t \leq t$ is also a random variable. Conditional on $M_t=m$, the distinct parents $\bm{Y}_1,...,\bm{Y}_m \stackrel{\mathrm{i.i.d}}{\sim} \sigma$. 

Let $\bm{Y}_j$ be a distinct parent from whose cluster an offspring is selected. Define normalized parent locations $\bm{U}_j:=\bm{Y}_j/\sqrt{d}$ for $1\leq j\leq M_t$. As mentioned, conditional on $M_t=m$, the $U_j$'s are i.i.d uniform on $\mathbb{S}^{d-1}$. We know that for two independent uniform points $\bm{U},\bm{V}\in \mathbb{S}^{d-1}$, we have $\mathbb E[\langle \bm{U},\bm{V}\rangle]=0$ and $E[\langle \bm{U},\bm{V}\rangle^2]=1/d$. Therefore, for every $\epsilon>0$, we get $\mathbb{P}\left(|\langle \bm{U},\bm{V}\rangle|>\epsilon\right)\leq\frac{1}{d\epsilon^2}\longrightarrow0$. 
Since $M_t\leq t$ and $t$ is fixed, a union bound gives
\begin{eqnarray}
	\max_{j\neq k}|\langle \bm{U}_j,\bm{U}_k\rangle|
\xrightarrow{P}0. \label{uv}
\end{eqnarray}
Now define $\bm{S}_d:=\sum_{j=1}^{M_t}\bm{U}_j$ and $\bm{V}_d:=\bm{S}_d / \|S_d\|$. For every $j$, we have
\begin{eqnarray}
	\langle \bm{S}_d,\bm{U}_j\rangle= 1+\sum_{k\neq j}\langle \bm{U}_k,\bm{U}_j\rangle.\label{un1}
\end{eqnarray}
Hence, by (\ref{uv}) and (\ref{un1}), we get $\max_{1\leq j\leq M_t}| \langle \bm{S}_d, \bm{U}_j\rangle -1 | \xrightarrow{P}0 $, which results into
\begin{eqnarray}
	\min_{1\leq j\leq M_t}\langle \bm{S}_d, \bm{U}_j \rangle = 1+o_P(1).
\end{eqnarray}
Furthermore, we know that $\|\bm{S}_d\|^2=\sum_{j=1}^{M_t}\|\bm{U}_j\|^2+2\sum_{1\leq j<k\leq M_t}
\langle \bm{U}_j,\bm{U}_k\rangle=M_t+o_{P}(1)$, resulting into $\|\bm{S}_d\|=\sqrt{M_t}+o_{P}(1)$. According to definition of $\bm{V}_d$ and by Slutsky's lemma, we obtain
\begin{eqnarray}
	\min_{1\leq j\leq M_t}\langle \bm{V}_d,\bm{U}_j\rangle
= \frac{1}{\sqrt{M_t}}+o_{P}(1).\label{Mt}
\end{eqnarray}
Since $M_t\leq t$, from (\ref{Mt}), it yields 
\begin{eqnarray}
	\min_{1\leq j\leq M_t}\langle \bm{V}_d,\bm{U}_j\rangle
\geq\frac1{\sqrt t}-o_{P}(1).
\end{eqnarray}
Now, by definition of geodesic distance in (\ref{geodis}), we can write
\begin{eqnarray}
	\max_{1\leq j\leq M_t}\frac{d_{geo}(\bm{Y}_j,\sqrt d\,\bm{V}_d)}{\sqrt d}\leq \arccos\frac1{\sqrt t}
+o_{P}(1).
\end{eqnarray}
Let $\bm{\Theta}_i$ be a chosen task optima with parent $\bm{Y}_j$, by triangle inequality and Proposition \ref{Prop1}, we have
\begin{eqnarray}
	\frac{d_{geo}(\sqrt d\,\bm{V}_d,\bm{\Theta}_i)}{\sqrt d} \leq \frac{d_{geo}(\sqrt d\,\bm{V}_d,\bm{Y}_j)}{\sqrt d}
+ \frac{d_{geo}(\bm{Y}_j,\bm{\Theta}_i)}{\sqrt d}\leq \arccos \frac{1}{\sqrt{t}}+ \arccos u_\alpha^\prime+o_{P}(1)
\end{eqnarray}
Since this holds uniformly over the fixed $1\leq i\leq t$, we get
\begin{eqnarray}
	\max_{1\leq i\leq t}\frac{d_{geo}(\sqrt d\,\bm{V}_d,\bm{\Theta}_i)}{\sqrt d}\leq \arccos \frac{1}{\sqrt{t}}+\arccos u_\alpha^\prime +o_{P}(1),
\end{eqnarray}
or for every $\epsilon >0$
\begin{eqnarray}
	\max_{1\leq i\leq t}d_{geo}(\sqrt d\,\bm{V}_d,\bm{\Theta}_i)
\leq\sqrt d\left(\arccos \frac{1}{\sqrt{t}}+\arccos u_\alpha^\prime +\epsilon\right), \label{or}
\end{eqnarray}
with high probability. For simplicity in notation, let 
\begin{eqnarray}
	r_\epsilon := \frac{r^*(d)}{\sqrt{d}}-\arccos u_\alpha^\prime -\arccos \frac{1}{\sqrt{t}} - \sqrt{d} \epsilon > 0,\label{reps}
\end{eqnarray}
where $r^*(d)/\sqrt{d} = \min_{1\leq i\leq t}r_i(d)/\sqrt{d}= \sqrt{2\eta/\lambda^*}$ and $\lambda^* = \max_{1\leq i\leq t}\lambda_i$ is a constant as in Lemma \ref{Lemma1}. 
Take any point $\bm{w}$ on geodesic ball $\bm{w}\in B_{\sqrt{d}\bm{V}_d}(r_\epsilon \sqrt{d})$. Again, by triangle inequality for every $i$, we get
\begin{eqnarray}
	d_{geo}(\bm{w},\bm{\Theta}_i) \leq d_{geo}(\bm{w},\sqrt{d}\bm{V}_d) + d_{geo}(\sqrt{d}\bm{V}_d, \bm{\Theta}_i) \stackrel{(a)}{\leq} r_\epsilon \sqrt{d}+ \sqrt{d}\left(\arccos \frac{1}{\sqrt{t}}+\arccos u_\alpha^\prime + \epsilon\right) = r^*(d),
\end{eqnarray}
where $(a)$ follows from (\ref{or}). Hence, $\bm{w}\in B_{\bm{\Theta}_i}(r^*(d))$ for every $i$ and therefore, 
\begin{eqnarray}
	B_{\sqrt{d}\bm{V}_d}(r_\epsilon \sqrt{d}) \subseteq \mathcal{K}_{t}(\bm{\Theta}_{1:t}),
\end{eqnarray}
with probability tending to $1$. 
Consequently,
\begin{eqnarray}
	\sigma\left(B_{\sqrt{d}\bm{V}_d}(r_\epsilon \sqrt{d}) \right) \leq \sigma(\mathcal{K}_t(\bm{\Theta}_{1:t})),\label{ballsub}
\end{eqnarray}
with high probability.
By our assumption that all $r_i(d)/\sqrt{d} < \pi/2$ and according to (\ref{reps}), we have $r_\epsilon < \pi/2$. According to \cite{Wyner1967}, for a geodesic ball $B_{\sqrt{d}\bm{V}_d}(r_\epsilon \sqrt{d})$ with $0<r_\epsilon <\pi/2$ and taking $\epsilon \downarrow 0$, we obtain
\begin{eqnarray}
	\lim_{d\rightarrow \infty}\frac{1}{d}\log \sigma \left(B_{\sqrt{d}\bm{V}_d}(r_\epsilon \sqrt{d})\right) = \log \sin \left(\sqrt{\frac{2\eta}{\lambda^*}} -\arccos u_\alpha^\prime -\arccos \frac{1}{\sqrt{t}}\right). \label{ballarea}
\end{eqnarray}
Combining (\ref{ballsub}) and (\ref{ballarea}), we have
\begin{eqnarray}
	\log \frac{1}{\sigma(\mathcal{K}_t(\bm{\Theta}_{1:t}))} < -d \log \sin \left(\sqrt{\frac{2\eta}{\lambda^*}} -\arccos u_\alpha^\prime -\arccos \frac{1}{\sqrt{t}}\right)+o(d),
\end{eqnarray}
with high probability.

For any fixed $\epsilon^\prime>0$, choose the budget $\rho_t = -d\log \sin \left(\sqrt{\frac{2\eta}{\lambda^*}} -\arccos u_\alpha^\prime -\arccos \frac{1}{\sqrt{t}}\right) + d\epsilon^\prime$. Then, for large $d$, with high probability,
\begin{eqnarray}
	\log \frac{1}{\nu_t} \leq \rho_t. 
\end{eqnarray}
As explained before, $\psi_{\rho_t}(\nu_t) = 1$ is achievable in this case, and therefore, $\mathbb{E}\left[\psi_{\rho_t}(\nu_t)\right] \rightarrow 1$. Thus, for any $\delta_t>0$, the chosen budget satisfies the constraint in (\ref{feasibility}) for large $d$. Finally, letting $\epsilon^\prime \rightarrow 0$, it yields
\begin{eqnarray}
	\lim_{d\rightarrow \infty}\frac{1}{d}C_{T}(\rho_{1:T}, \delta_{1:T}, d) \leq \sum_{t=1}^{T}-\log \sin\left(\sqrt{\frac{2\eta}{\lambda^*}} -\arccos u_\alpha^\prime -\arccos \frac{1}{\sqrt{t}}\right).
\end{eqnarray}

\section{Conclusion}
We analyzed the effect of spherical parameter space on the feasible region of continual learning, under the assumption that task optima distributed by Neyman-Scott clustered point process $\Phi$. We derived an upper bound, dependent on the maximum eigenvalue of task Hessians, on the number of sequential tasks, guaranteeing the existence of a nonempty feasible region asymptotically. We further look at the continual learning information theoretically, modeling the learning process of each task as a channel $Q_t$ and the whole process of continual learning as sequence of channels. To  derive a bounded capacity for continual learning, we constrained the KL divergence between the distribution of each task $t$ parameter $\bm{W}_t$ and the initialization distribution $\sigma$ by budget $\rho_t$. Using this, we define the channel capacity for a fixed $d$ as the mutual information between the set of sampled task optima from $\Phi$ and the set of outputs of channels under the constraint of non-forgetting. We analyzed mathematically the maximum mass that each channel $Q_t$ can put on their corresponding feasible region $\mathcal{K}_{t}(\bm{\theta}_{1:t})$. We also derived the explicit relation between $\rho_t$ and log normalized volume of feasible region for simple channels whose output is a uniformly distributed point on $\mathcal{K}_t(\bm{\theta}_{1:t})$.

\appendices
\section{Proof of Lemma \ref{Lemma1}}\label{ProofLemma1}
\begin{proof}
We want to approximate $R_i(\bm w)$ near $\bm{\Theta}_i$. However, because $\bm w$ lies on the sphere, we cannot directly perform a standard Taylor expansion with respect to $\bm w$. To fix this, we define the map $E_{\bm{\Theta}_i}(\bm v): \mathcal{T}_{\bm{\Theta}_i} \longrightarrow
\mathbb{S}^{d-1}(\sqrt{d})$ as
\begin{eqnarray}
	E_{\bm{\Theta}_i}(\bm v)= \bm{\Theta}_i \cos\frac{\|\bm v\|}{\sqrt d} + \sqrt d \frac{\bm v}{\|\bm v\|} \sin\frac{\|\bm v\|}{\sqrt d}, \qquad
\forall \bm v\in\mathcal{T}_{\bm{\Theta}_i}.
\label{E}
\end{eqnarray}
This map gives the point on the sphere reached by moving from $\bm{\Theta}_i$ along a geodesic with initial tangent vector $\bm v$. We can therefore write $f_i(\bm v)=R_i\left(E_{\bm{\Theta}_i}(\bm v)\right)$, where $f_i(\bm v)$ is the loss function expressed in the local geodesic coordinates around $\bm{\Theta}_i$.
Since $ E_{\bm{\Theta}_i}(\bm 0)=\bm{\Theta}_i$, 
and $\bm{\Theta}_i$ is a local minimizer of $R_i$, we have $f_i(\bm 0)=R_i(\bm{\Theta}_i)$ and 
$\nabla f_i(\bm 0)=\bm 0$.

Let $\mathbf H_i$ denote the Hessian matrix of $f_i$ at $\bm 0$, viewed as a basis of $\mathcal{T}_{\bm{\Theta}_i}$. Thus,
\begin{eqnarray}
\bm v^{\top}\mathbf H_i\bm v= \sum_{n,m=1}^{d-1} v_nv_m \left( \left. \frac{\partial^2 f_i} {\partial v_n\partial v_m} \right|_{\bm v=\bm 0} \right).
\end{eqnarray}
By the isotropy assumption, we get $\bm v^{\top}\mathbf H_i\bm v = \lambda_i\|\bm v\|^2$.

We assume that the loss function $f_i$ admits the following asymptotic quadratic approximation 
\begin{eqnarray}
	\frac{1}{d} \left( f_i(\sqrt d\bm z)-f_i(\bm 0)
\right)= \frac{\lambda_i}{2}\|\bm z\|^2 + o(1),
\label{Taylor}
\end{eqnarray}
uniformly for $\|\bm z\|$. Equivalently,
\begin{eqnarray}
	f_i(\sqrt d\bm z)= f_i(\bm 0) + d\frac{\lambda_i}{2}\|\bm z\|^2 + o(d).
\label{Taylor2}
\end{eqnarray}
Since $f_i(\bm 0)=R_i(\bm{\Theta}_i)$, we obtain
\begin{eqnarray}
R_i\left(E_{\bm{\Theta}_i}(\sqrt d\bm z)\right)=
R_i(\bm{\Theta}_i) + d\frac{\lambda_i}{2}\|\bm z \|^2 + o(d).
\label{Taylor3}
\end{eqnarray}

The acceptable loss condition is $R_i(\bm w)-R_i(\bm{\Theta}_i) \leq \eta d$. Setting $\bm w = E_{\bm{\Theta}_i}(\sqrt d \ \bm z)$ and using (\ref{Taylor3}), we obtain
\begin{eqnarray}
	d\frac{\lambda_i}{2}\|\bm z\|^2+o(d)\leq \eta d.
\end{eqnarray}
Dividing by $d$ gives
\begin{eqnarray}
	\frac{\lambda_i}{2}\|\bm z\|^2+o(1)\leq\eta.
\end{eqnarray}
Hence, by $\bm v=\sqrt d\,\bm z$, we have
\begin{eqnarray}
	\|\bm v\| \leq \sqrt{\frac{2\eta}{\lambda_i}} \sqrt d + o(\sqrt d).
\label{v}
\end{eqnarray}
On the other hand, by the definition of the map in (\ref{E}), the geodesic distance from $E_{\bm{\Theta}_i}(\bm v)$ to $\bm{\Theta}_i$ is exactly $d\left( E_{\bm{\Theta}_i}(\bm v), \bm{\Theta}_i \right) = \|\bm v\|$. Therefore, combining this identity with (\ref{v}), the acceptable loss region 
\begin{eqnarray}
	\mathcal{L}_i \rightarrow \left\{ \bm w\in\mathbb{S}^{d-1}(\sqrt{d}): d_{geo}(\bm w,\bm{\Theta}_i) \leq
\sqrt{\frac{2\eta}{\lambda_i}}\sqrt d + o(\sqrt d) \right\},
\end{eqnarray}
as $d\rightarrow \infty$.

\begin{remark}
	We restrict the uniform approximation condition to every fixed $C<\pi$, which is the normalized injectivity radius of the sphere. The antipodal point $-\bm\Theta_i$ is reached at geodesic distance $\pi\sqrt d$, since $E_{\bm\Theta_i}(\pi\sqrt d\bm u)= -\bm\Theta_i$ for every unit vector $\bm u\in\mathcal T_{\bm\Theta_i}$. Hence, the map $E_{\bm \Theta_i}$ ceases to be injective at radius $\pi\sqrt d$, and $-\bm\Theta_i$ is the cut-locus point of $\bm\Theta_i$. Consequently, in the normalized coordinates $\bm z=\bm v/\sqrt d$, the geodesic coordinate chart is one-to-one only for $\|\bm z\|<\pi$.
Thus, the uniform approximation assumption is stated for every fixed $C<\pi$, ensuring that the supremum is taken over a compact subset of the injectivity domain.
\end{remark}
\end{proof}

\section{Proof of Proposition \ref{Prop1}} \label{ProofProp1}

\begin{proof}
Let $U:=\langle \bm{Y}_j,\bm{\Theta}_i\rangle$. Given $N_p$, the parent point $\bm{Y}_j$ is uniformly distributed on $\mathbb{S}^{d-1}(\sqrt{d})$, while, conditional on $\bm{Y}_j$, the offspring point $\bm{\Theta}_i$ is sampled according to $\mathrm{vMF}(\bm{Y}_j,\kappa)$.
We derive the distribution of $U$.

Given a realization $\bm{\Theta}_i=\bm{\theta}_i$, the conditional density of $\bm{Y}_j$ follows from Bayes' rule. Indeed, since $\bm{Y}_j\sim\mathrm{Uniform}(\mathbb{S}^{d-1}(\sqrt{d}))$, we have
\begin{eqnarray}
	f_{\bm{Y}_j}(\bm{y}_j)=\frac{1}{|\mathbb{S}^{d-1}(\sqrt{d})|},
\qquad\bm{y}_j\in\mathbb{S}^{d-1}(\sqrt{d}),
\end{eqnarray}
and, by the described Neyman-Scott cluster point process model, 
\begin{eqnarray}
	f_{\bm{\Theta}_i\mid\bm{Y}_j}(\bm{\theta}_i\mid\bm{y}_j)
=\frac{C_d(\kappa)}{d^{(d-1)/2}}
\exp\left\{\frac{\kappa\langle \bm{y}_j,\bm{\theta}_i\rangle}{d}\right\}.
\end{eqnarray}
Therefore, the joint density is 
\begin{eqnarray}
	f_{\bm{\Theta}_i,\bm{Y}_j}(\bm{\theta}_i,\bm{y}_j) =f_{\bm{\Theta}_i\mid\bm{Y}_j}(\bm{\theta}_i\mid\bm{y}_j)
f_{\bm{Y}_j}(\bm{y}_j)  =\frac{C_d(\kappa)}
{d^{(d-1)/2}|\mathbb{S}^{d-1}(\sqrt{d})|}
\exp\left\{\frac{\kappa\langle \bm{y}_j,\bm{\theta}_i\rangle}{d}\right\}.
\label{Joint}
\end{eqnarray}
We first derive the marginal distribution of $\bm{\Theta}_i$. By integrating \eqref{Joint} over $\mathbb{S}^{d-1}(\sqrt{d})$, it yields
\begin{eqnarray}
	f_{\bm{\Theta}_i}(\bm{\theta}_i) = \int_{\mathbb{S}^{d-1}(\sqrt{d})}
f_{\bm{\Theta}_i,\bm{Y}_j}(\bm{\theta}_i,\bm{y}_j)\mathrm{d}\sigma(\bm{y}_j)=\frac{C_d(\kappa)}
{d^{(d-1)/2}|\mathbb{S}^{d-1}(\sqrt{d})|}
\int_{\mathbb{S}^{d-1}(\sqrt{d})}
\exp\left\{\frac{\kappa\langle \bm{y}_j,\bm{\theta}_i\rangle}{d}\right\}
\mathrm{d}\sigma(\bm{y}_j).
\label{ThetaMarginal}
\end{eqnarray}
For every fixed $\bm{\theta}_i\in\mathbb{S}^{d-1}(\sqrt{d})$, the term
\begin{eqnarray}
	\frac{C_d(\kappa)}{d^{(d-1)/2}} \exp\left\{\frac{\kappa\langle \bm{y}_j,\bm{\theta}_i\rangle}{d}\right\}
\end{eqnarray}
is the density of a $\mathrm{vMF}(\bm{\theta}_i,\kappa)$ distribution with
respect to the surface measure (Haar measure) $\sigma$. Hence, by the fact that the integral of  $\mathrm{vMF}$ density over the whole space is $1$, we obtain
\begin{eqnarray}
	\int_{\mathbb{S}^{d-1}(\sqrt{d})}
\exp\left\{\frac{\kappa\langle \bm{y}_j,\bm{\theta}_i\rangle}{d}\right\}\mathrm{d}\sigma(\bm{y}_j)=\frac{d^{(d-1)/2}}{C_d(\kappa)}.
\label{vMFIntegral}
\end{eqnarray}

Substituting \eqref{vMFIntegral} into \eqref{ThetaMarginal} gives
\begin{eqnarray}
	f_{\bm{\Theta}_i}(\bm{\theta}_i)=\frac{C_d(\kappa)}{d^{(d-1)/2}|\mathbb{S}^{d-1}(\sqrt{d})|}\frac{d^{(d-1)/2}}{C_d(\kappa)}
=\frac{1}{|\mathbb{S}^{d-1}(\sqrt{d})|}.
\label{ThetaUniform}
\end{eqnarray}

Next, by Bayes' rule, the conditional density of $\bm{Y}_j$ given
$\bm{\Theta}_i=\bm{\theta}_i$ is
\begin{eqnarray}
	f_{\bm{Y}_j\mid\bm{\Theta}_i}(\bm{y}_j\mid\bm{\theta}_i)=\frac{f_{\bm{\Theta}_i,\bm{Y}_j}
(\bm{\theta}_i,\bm{y}_j)}
{f_{\bm{\Theta}_i}(\bm{\theta}_i)}=\frac{ \frac{C_d(\kappa)}{d^{(d-1)/2}|\mathbb{S}^{d-1}(\sqrt{d})|}\exp\left\{\frac{\kappa\langle \bm{y}_j,\bm{\theta}_i\rangle}{d}\right\}}
{\frac{1}{|\mathbb{S}^{d-1}(\sqrt{d})|}}=\frac{C_d(\kappa)}{d^{(d-1)/2}}\exp\left\{
\frac{\kappa\langle \bm{y}_j,\bm{\theta}_i \rangle}{d}\right\},
\label{YgivenTheta}
\end{eqnarray}
resulting into $\bm{Y}_j\mid\bm{\Theta}_i=\bm{\theta}_i\sim\mathrm{vMF}(\bm{\theta}_i,\kappa)$.

We now derive the conditional distribution of $U=\langle \bm{Y}_j,\bm{\Theta}_i\rangle$ given $\bm{\Theta}_i=\bm{\theta}_i$. Fix $\bm{\theta}_i\in\mathbb{S}^{d-1}(\sqrt{d})$ and let $Q_{\bm{\theta}_i}$ be an orthogonal matrix satisfying $Q_{\bm{\theta}_i}\bm{\theta}_i=\sqrt{d}\bm{e}_1$. Let $\bm{Z}_j:=Q_{\bm{\theta}_i}\bm{Y}_j$.
Since $Q_{\bm{\theta}_i}$ is orthogonal, then we have
\begin{eqnarray}
	U=\langle \bm{Y}_j,\bm{\theta}_i\rangle=\left\langle Q_{\bm{\theta}_i}\bm{Y}_j Q_{\bm{\theta}_i}\bm{\theta}_i\right\rangle =\langle \bm{Z}_j,\sqrt{d}\bm{e}_1\rangle =\sqrt{d}Z_{j,1},
\label{Rotation}
\end{eqnarray}
where $Z_{j,1}$ is the first entry of $\bm{Z}_j$.
Moreover, by \eqref{YgivenTheta}, we have
\begin{eqnarray}
	f_{\bm{Z}_j\mid\bm{\Theta}_i}(\bm{z}_j\mid\bm{\theta}_i)
=\frac{C_d(\kappa)}{d^{(d-1)/2}}
\exp\left\{\frac{\kappa z_{j,1}}{\sqrt{d}}\right\}, \qquad \forall \bm{z}_j\in\mathbb{S}^{d-1}(\sqrt{d}).
\label{Zdensity}
\end{eqnarray}
For $-d\leq u\leq d$, the conditional cumulative distribution function of $U=\sqrt{d},Z_{j,1}$ according to \eqref{Zdensity} is
\begin{eqnarray}
	&& F_{U\mid\bm{\Theta}_i}(u \vert\bm{\theta}_i)
=\mathbb{P}\left[Z_{j,1}\leq \frac{u}{\sqrt{d}}
\middle \vert \bm{\Theta}_i=\bm{\theta}_i\right] = \int_{{\bm z_j\in\mathbb S^{d-1}(\sqrt{d}):
z_{j,1}\leq u/\sqrt{d}}}f_{\bm Z_j\mid\bm{\Theta}_i}(\bm z_j\mid\bm{\theta}_i)\mathrm d\sigma(\bm z_j) \nonumber \\
	&& \ \ \ \ \ \ \ \ \ \ \ \ \ \ \ =\frac{C_d(\kappa)}{d^{(d-1)/2}}
\int_{{\bm z_j\in\mathbb S^{d-1}(\sqrt{d}):
z_{j,1}\leq u/\sqrt{d}}} e^{\kappa z_{j,1}/\sqrt{d}}\mathrm d\sigma(\bm z_j).
\label{UCDF1}
\end{eqnarray}
To evaluate the integral in (\ref{UCDF1}), we follow the parameterization of $\mathbb S^{d-1}(\sqrt{d})$ by $\bm z_j =\left(\sqrt{d}\cos\phi,\sqrt{d}\sin\phi\bm v\right)$, where
$0\leq\phi\leq\pi$ and $\bm v\in\mathbb S^{d-2}$.
Under this parameterization, the surface area element is $\mathrm d\sigma(\bm z_j)=d^{(d-1)/2}(\sin\phi)^{d-2}\mathrm d\phi\mathrm{d} \sigma(\bm v)$. 
Moreover, $z_{j,1}=\sqrt{d}\cos\phi$.
Thus, the event ${z_{j,1}\leq u/\sqrt{d}}$ is equivalent to
\begin{eqnarray}
	\left\{\sqrt{d}\cos\phi\leq\frac{u}{\sqrt{d}}\right\}=\left\{\phi\geq\arccos\left(\frac{u}{d}\right)\right\}.
\end{eqnarray}
Consequently, by \eqref{UCDF1}, we have
\begin{eqnarray}
	&& F_{U\mid\bm{\Theta}_i}(u\mid\bm{\theta}_i)= C_d(\kappa)\int_{\arccos(u/d)}^{\pi}
\int_{\mathbb S^{d-2}}e^{\kappa\cos\phi}(\sin\phi)^{d-2}\mathrm d\sigma(\bm v)\mathrm d\phi \nonumber\\
	&& \ \ \ \ \ \ \ \ \ \ \ \ \ \ \ \ \ =C_d(\kappa)|\mathbb S^{d-2}|\int_{\arccos(u/d)}^{\pi}e^{\kappa\cos\phi}
(\sin\phi)^{d-2}\mathrm d\phi.
\label{UCDF2}
\end{eqnarray}
By a simple change of variable $t=\cos\phi$, we rewrite (\ref{UCDF2}) as
\begin{eqnarray}
	F_{U\mid\bm{\Theta}_i}(u\mid\bm{\theta}_i)= C_d(\kappa)|\mathbb S^{d-2}|
\int_{-1}^{u/d}e^{\kappa t}(1-t^2)^{\frac{d-3}{2}}\mathrm dt.
\label{UCDF3}
\end{eqnarray}
Since the integrand in \eqref{UCDF3} is continuous for $-1<t<1$, the conditional density of $U$ is obtained by differentiating (\ref{UCDF3}) with respect to $u$ as
\begin{eqnarray}
	f_{U\mid\bm{\Theta}_i}(u\mid\bm{\theta}_i)=
\frac{C_d(\kappa)|\mathbb S^{d-2}|}{d}
e^{\kappa u/d}\left(1-\frac{u^2}{d^2}\right)^{\frac{d-3}{2}}, \qquad -d<u<d.
\label{UgivenThetaCorrect}
\end{eqnarray}
Note that the right-hand side of (\ref{UgivenThetaCorrect}) does not depend on $\bm{\theta}_i$.
By \eqref{ThetaUniform}, the density of $U$ for $|u|<d$ is
\begin{eqnarray}
	f_U(u)= \frac{C_d(\kappa)|\mathbb S^{d-2}|}{d}e^{\kappa u/d}
\left(1-\frac{u^2}{d^2}\right)^{\frac{d-3}{2}},
\label{UfinalCompact}
\end{eqnarray}
where $|\mathbb{S}^{d-2}|$ is the surface area of unit sphere in $\mathbb{R}^{d-1}$ which equals
\begin{eqnarray}
|\mathbb S^{d-2}|=\frac{2\pi^{(d-1)/2}}
{\Gamma\left(\frac{d-1}{2}\right)},
\end{eqnarray}
and $C_d(\kappa)$ is given in (\ref{Cd}). Note that $f_U(u)=0$ for $|u|>d$. Now, let $U^\prime = U/d$, then from (\ref{UfinalCompact}), we obtain
\begin{eqnarray}
	f_{U^\prime}(u^\prime) = C_d(\kappa)|\mathbb{S}^{d-2}|e^{\kappa u^{\prime}}(1-{u^\prime}^2)^{\frac{d-3}{2}}, 
\end{eqnarray}
for $|u^\prime|<1$ and $f_{U^\prime}(u^\prime)=0$ for $|u^\prime|>1$. We rewrite the density of $U^\prime$ as 
\begin{eqnarray}
	f_{U^\prime}(u^\prime) = C_d(\kappa)|\mathbb{S}^{d-2}| e^{L_d(u^\prime)}, \qquad |u^\prime| <1,
\end{eqnarray}
where
\begin{eqnarray}
	L_d(u^\prime)=\frac{d-3}{2}\log (1-{u^\prime}^2)+\kappa u^\prime.
\end{eqnarray}
Since $\kappa /d \rightarrow \alpha$, for a fixed $u^\prime \in (-1,1)$, we have
\begin{eqnarray}
	\frac{1}{d}L_{d}(u^\prime) = \frac{d-3}{2d}\log (1-{u^\prime}^2)+\frac{\kappa}{d} u^\prime \longrightarrow \frac{1}{2}\log (1-{u^\prime}^2) + \alpha u^\prime =: \ell_{\alpha}(u^\prime),
\end{eqnarray}
as $d\rightarrow \infty$. 

It is simple to see that for any interval $\mathcal{I} \subset (-1,1)$, we get
\begin{eqnarray}
	\mathbb{P}\left[U^\prime \in \mathcal{I}\right] = \frac{\int_{\mathcal{I}}e^{L_d(u^\prime)}\mathrm{d}u^\prime}{\int_{-1}^{1}e^{L_d(u^\prime)}\mathrm{d}u^\prime}. \label{prob}
\end{eqnarray}
Since $L_d(u^\prime)/d \rightarrow \ell_{\alpha}(u^\prime)$ uniformly on every compact subsets of $(-1,1)$, and since $\ell_\alpha(u^\prime)\rightarrow -\infty$ as $u^\prime \rightarrow \pm 1$, the contribution from neighborhoods of boundary $(-1,1)$ is exponentially negligible. Hence, by Laplace's method \cite{MacKay2003}, we have
\begin{eqnarray}
	\frac{1}{d}\log \int_{-1}^{1}e^{L_d(u^\prime)}\mathrm{d}u^\prime \longrightarrow \max_{-1<u^\prime<1} \ell_\alpha(u^\prime), \label{limit1}
\end{eqnarray}
as $d\rightarrow \infty$.
Taking derivative of $\ell_\alpha(u^\prime)$ and setting it to zero yields
\begin{eqnarray}
	\frac{\mathrm{d}\ell_\alpha(u^\prime)}{\mathrm{d}u^\prime} = \frac{-u^\prime}{1-{u^\prime}^2}+\alpha =0.\label{deri}
\end{eqnarray}
Solving for $u^\prime$ in (\ref{deri}), we get
\begin{eqnarray}
	u_{\alpha}^{\prime} = \frac{\sqrt{1+4\alpha^2}-1}{2\alpha}.\label{ualpha}
\end{eqnarray}
Since 
\begin{eqnarray}
	\frac{\mathrm{d}^2\ell_\alpha(u^\prime)}{\mathrm{d}{u^\prime}^2} = -\frac{1+{u^\prime}^2}{(1-{u^\prime}^2)^2}<0,
\end{eqnarray}
and $u_{\alpha}^\prime\in (-1,1)$, then $u_\alpha^\prime$ is the max of $\ell_\alpha(u^\prime)$ on $(-1,1)$. For a $\delta>0$, let $\mathcal{I}_{\delta}:= (-1,1) \backslash (u_\alpha^\prime -\delta, u_\alpha^\prime + \delta)$. Define $\Delta_\delta:= \ell_\alpha(u_\alpha^\prime) - \sup_{u^\prime\in \mathcal{I}_{\delta}} \ell_{\alpha}(u^\prime) >0$. Then, similarly, by Laplace's method, we have
\begin{eqnarray}
	\frac{1}{d}\log \int_{\mathcal{I}_{\delta}} e^{L_d(u^\prime)}\mathrm{d}u^\prime \longrightarrow \sup_{u^\prime \in \mathcal{I}_{\delta}} \ell_{\alpha}(u^\prime),\label{limit2}
\end{eqnarray}
as $d\rightarrow \infty$.
Therefore, by (\ref{prob}), (\ref{limit1}), (\ref{ualpha}), and (\ref{limit2}), we get
\begin{eqnarray}
	\lim_{d\rightarrow \infty}\mathbb{P}\left[U^\prime \in \mathcal{I}_{\delta}\right] = \lim_{d\rightarrow \infty}\mathbb{P}\left[|U^\prime -u_\alpha| > \delta \right] = \lim_{d\rightarrow}e^{-d\Delta_{\delta}} = 0.\label{converg}
\end{eqnarray}
Hence, by definition of convergence in probability, from (\ref{converg}), we have
\begin{eqnarray}
	\frac{\langle \bm{Y}_j, \bm{\Theta}_i \rangle}{d} \xrightarrow{P} u_{\alpha}^\prime. 
\end{eqnarray}
Since $\mathbb{P}[u_\alpha^\prime \in (-1,1)] = 1$, by continuous mapping theorem, we obtain
\begin{eqnarray}
	\arccos \left(\frac{\langle \bm{Y}_j, \bm{\Theta}_i\rangle}{d}\right) \xrightarrow{P} \arccos u_{\alpha}^\prime. \label{arcos}
\end{eqnarray}
Finally, comparing the result from (\ref{arcos}) and the definition of geodesic distance in (\ref{geodis}) completes the proof. 
\end{proof}

\section{Proof of Theorem \ref{Thm1}} \label{ProofThm1}
\begin{proof}
We sample $T$ points uniformly without replacement from $\Phi$. Let $\{ \bm{\Theta}_1, ..., \bm{\Theta}_T\}$, be the set of sampled points (task optima). For each sampled point $\bm{\Theta}_i$, let $J_i$ be the index of its parent. That is, if $\bm{\Theta}_i$ lies in cluster $j$, then $J_i = j$. As before, we denote the parent of $\bm{\Theta}_i$ by $\bm{Y}_{J_i}$. We assume that radii of acceptable loss regions are large enough such that for a fixed $d$,
\begin{eqnarray}
	\frac{r^*(d)}{\sqrt{d}}:= \frac{1}{\sqrt{d}}\min_{1\leq i \leq T} {r_i(d)} > \arccos u^\prime_{\alpha},
\end{eqnarray} 
or such that for a fixed $\epsilon >0$, we have $r^*(d)/\sqrt{d} - \arccos u^\prime_\alpha - \epsilon >0$. Note that as shown in Lemma \ref{Lemma1}, the ratio $r_i(d)/\sqrt{d}$ is a constant and does not depend on $d$.

We define the event 
\begin{eqnarray}
	\mathcal{A}_{\epsilon}:= \left\{ \frac{1}{\sqrt{d}}\max_{1\leq i\leq T}{d_{geo}(\bm{\Theta}_i,\bm{Y}_{J_i})} \leq \arccos u^\prime_\alpha + \epsilon \right \}.\label{A}
\end{eqnarray}
Here, the event $\mathcal{A}_\epsilon$ says that every offspring point is close to its parent. We also define the other event
\begin{eqnarray}
	\mathcal{B}_\epsilon :=\left\{\exists \bm{w}\in \mathbb{S}^{d-1}(\sqrt{d}) \ \mathrm{s.t.}\ \frac{1}{\sqrt{d}}\max_{j\in \mathcal{S}}d_{geo}(\bm{w},\bm{Y}_j) \leq \frac{r^*(d)}{\sqrt{d}}- \arccos u^\prime_\alpha -\epsilon \right\},\label{B}
\end{eqnarray}
where $\mathcal{S}=\{ J_{i}: 1\leq i\leq T\}$ is the set of distinct parent labels. The event $\mathcal{B}_\epsilon$ says that the distinct parent points can be covered by a geodesic ball of radius $r^*(d)/\sqrt{d}-\arccos u^\prime_\alpha -\epsilon$.

Suppose $\mathcal{A}_\epsilon$ and $\mathcal{B}_\epsilon$ both hold. Since $\mathcal{B}_\epsilon$ holds, then there exists a point on sphere $\bm{w} \in \mathbb{S}^{d-1}(\sqrt{d})$ such that for every parent point $\bm{Y}_{J_i}$, we have
\begin{eqnarray}
	\frac{d_{geo}(\bm{w},\bm{Y}_{J_i})}{\sqrt{d}} \leq \frac{r^*(d)}{\sqrt{d}}-\arccos u^\prime_\alpha-\epsilon.\label{ro1}
\end{eqnarray}
Since $\mathcal{A}_\epsilon$ holds, for every offspring point $\bm{\Theta}_i$, we have
\begin{eqnarray}
	\frac{d_{geo}(\bm{\Theta}_i,\bm{Y}_{J_i})}{\sqrt{d}} \leq \arccos u^\prime_\alpha + \epsilon.\label{ro2}
\end{eqnarray}
By triangle inequality, we obtain
\begin{eqnarray}
	\frac{1}{\sqrt{d}}d_{geo}(\bm{w},\bm{\Theta}_i)\leq \frac{1}{\sqrt{d}}d_{geo}(\bm{w},\bm{Y}_{J_i})+\frac{1}{\sqrt{d}}d_{geo}(\bm{Y}_{J_i},\bm{\Theta}_i) \stackrel{(b)}{\leq} \frac{r^*(d)}{\sqrt{d}},
\end{eqnarray}
where $(b)$ follows from (\ref{ro1}) and (\ref{ro2}).
But, $r^*(d) \leq r_i(d)$ for all $i\in\{1,...,T\}$. Hence, 
\begin{eqnarray}
	d_{geo}(\bm{w},\bm{\Theta}_i) \leq {r_i(d)}, \qquad \forall i\in\{1,...,T\}.
\end{eqnarray}
Thus, 
\begin{eqnarray}
	\bm{w}\in \mathcal{K}_{T}(\bm{\Theta}_{1:T})=\bigcap_{i=1}^{T}B_{\bm{\Theta}_i}(r_i(d)).
\end{eqnarray}
Therefore, 
\begin{eqnarray}
	\mathcal{A}_{\epsilon}\cap \mathcal{B}_{\epsilon} \subseteq \left\{\mathcal{K}_{T}(\bm{\Theta}_{1:T}) \neq \emptyset \right\}.\label{cap}
\end{eqnarray}
Simply by applying the probability measure to the both sides of (\ref{cap}) 
\begin{eqnarray}
	1-\mathbb{P}\left[\mathcal{A}_{\epsilon}^{c}\right]-\mathbb{P}\left[\mathcal{B}_{\epsilon}^{c}\right]\stackrel{(c)}{\leq} \mathbb{P}\left[\mathcal{A}_{\epsilon}\cap \mathcal{B}_{\epsilon} \right] \leq \mathbb{P}\left[\mathcal{K}_{T}(\bm{\Theta}_{1:T}) \neq \emptyset\right],
\end{eqnarray}
where $(c)$ is by union bound. We next prove that both probabilities $\mathbb{P}\left[\mathcal{A}_{\epsilon}^{c}\right]$ and $\mathbb{P}\left[\mathcal{B}_{\epsilon}^{c}\right]$ tend to $0$ under an upper bound on $T$ as $d\rightarrow \infty$.
\subsection{Proving $\mathbb{P}\left[\mathcal{A}_{\epsilon}^{c}\right] \rightarrow 0$ as $d\rightarrow \infty$:}

By Proposition \ref{Prop1}, we proved that given $N_p$, the normalized geodesic distance between a parent $\bm{Y}_{J_i}$ and one of its offspring point $\bm{\Theta}_i$ converges in probability to $\arccos u^\prime_\alpha$ when $\kappa/d \rightarrow \alpha$. Hence, since 
\begin{eqnarray}
	\left\{\frac{1}{\sqrt{d}}d_{geo}(\bm{\Theta}_i,\bm{Y}_{J_i}) > \arccos u^\prime_\alpha +\epsilon\right\} \subset \left\{\left | \frac{1}{\sqrt{d}} d_{geo}(\bm{\Theta}_i,\bm{Y}_{J_i})-\arccos u^\prime_\alpha \right | > \epsilon \right\},
\end{eqnarray}
we obtain
\begin{eqnarray}
	\mathbb{P}\left[\frac{1}{\sqrt{d}}d_{geo}(\bm{\Theta}_i,\bm{Y}_{J_i}) > \arccos u^\prime_\alpha +\epsilon \right]\rightarrow 0,\label{ConvePRob}
\end{eqnarray}
as $d\rightarrow \infty$.
By union bound and according to the definition of event $\mathcal{A}_\epsilon$ in (\ref{A}), we have
\begin{eqnarray}
	\mathbb{P}\left[\mathcal{A}_{\epsilon}^{c}\right]= \mathbb{P}\left[\exists i\in \{1,...,T\}: \frac{1}{\sqrt{d}}d_{geo}(\bm{\Theta}_i,\bm{Y}_{J_i}) > \arccos u^\prime_\alpha + \epsilon\right]\leq \sum_{i=1}^{T}\mathbb{P}\left[\frac{1}{\sqrt{d}}d_{geo}(\bm{\Theta}_i,\bm{Y}_{J_i}) > \arccos u^\prime_\alpha + \epsilon\right]\rightarrow 0,\label{AcompleZero}
\end{eqnarray}
where (\ref{AcompleZero}) follows from (\ref{ConvePRob}) and $T$ being a constant. Note that here $\mathbb{P}[\mathcal{A}_{\epsilon}^{c}]$ goes to zero regardless of $T$.

\subsection{Proving $\mathbb{P}\left[\mathcal{B}_{\epsilon}^{c}\right]\rightarrow 0$ as $d\rightarrow \infty$:}

The number of distinct parents $|\mathcal{S}|$ is a random variable satisfying $|\mathcal{S}| \leq T$ almost surely. We prove that there is a deterministic threshold $N$ such that if $T \leq N$, then $\mathbb{P}\left[\mathcal{B}_{\epsilon}^{c}\right]\rightarrow 0$ as $d\rightarrow \infty$. Before proceeding, let the minimal covering radius of distinct parents be
\begin{eqnarray}
	r_{|\mathcal{S}|}(d):= \min_{\bm{w}\in \mathbb{S}^{d-1}(\sqrt{d})} \max_{j\in \mathcal{S}} \ {d_{geo}(\bm{w},\bm{Y}_{j})}.
\end{eqnarray}
Then, event $\mathcal{B}_{\epsilon}$ can be rewritten as 
\begin{eqnarray}
	\mathcal{B}_{\epsilon}=\left \{ \frac{r_{|\mathcal{S}|}(d)}{\sqrt{d}} \leq \frac{r^*(d)}{\sqrt{d}}-\arccos u^\prime_\alpha-\epsilon\right \}.\label{Bcomple}
\end{eqnarray}
The radius $r_{| \mathcal{S}|}(d)$ is defined as the minimal possible radius of a spherical cap (geodesic ball) on $\mathbb{S}^{d-1}(\sqrt{d})$ that contains all parents in $\mathcal{S}$. If we add more parents, meaning increasing $| \mathcal{S}|$, then $r_{|\mathcal{S}|}$ must increase as well, i.e., for a realization of $| \mathcal{S}| = s$ and a constant integer $n$, if $s\leq n$, then for all $\bm{w}\in \mathbb{S}^{d-1}(\sqrt{d})$, we have $r_s(d) \leq r_n(d)$. Therefore, 
\begin{eqnarray}
	r_{| \mathcal{S}| }(d)\stackrel{(f)}{\leq} r_{T}(d) \leq r_N(d),\label{r}
\end{eqnarray}
where $(f)$ is an almost sure inequality. Therefore, by (\ref{Bcomple}) and (\ref{r}), we get
\begin{eqnarray}
	\mathcal{B}_{\epsilon}^{c} = \left \{ \frac{r_{|\mathcal{S}|}(d)}{\sqrt{d}} > \frac{r^*(d)}{\sqrt{d}}-\arccos u^\prime_\alpha-\epsilon\right \} \subseteq \left\{\frac{r_N(d)}{\sqrt{d}} >\frac{r^*(d)}{\sqrt{d}}-\arccos u^\prime_\alpha-\epsilon \right\},
\end{eqnarray}
resulting into
\begin{equation}
	\mathbb{P}\left[\mathcal{B}_{\epsilon}^{c}\right] \leq \mathbb{P}\left[\frac{r_N(d)}{\sqrt{d}} >\frac{r^*(d)}{\sqrt{d}}-\arccos u^\prime_\alpha-\epsilon\right]. \label{Prim}
\end{equation}

Let $\bm{Z}_1, ..., \bm{Z}_N$ be i.i.d Gaussian vectors in $\mathbb{R}^{d}$ i.e., $\bm{Z}_j\sim \mathcal{N}(\bm{0},\mathbf{I}_d)$. Define $\bm{Y}_j = \sqrt{d}\bm{Z}_j/\| \bm{Z}_j\|$. Then $\bm{Y}_j$ is uniformly distributed on $\mathbb{S}^{d-1}(\sqrt{d})$. Define the Gram matrix 
\begin{eqnarray}
	\mathbf{G} := \frac{(\langle \bm{Y}_j,\bm{Y}_k\rangle)_{1\leq j,k\leq N}}{d},
\end{eqnarray}
whose entries are $G_{jk}=\langle \bm{Y}_j,\bm{Y}_k\rangle/d$ for $1\leq j,k\leq N$. 
Let $\mathbf{Y}$ be a $d\times N$ matrix with $\bm{Y}_j$ as its columns. Then we can write the Gram matrix $\mathbf{G}$ as $\mathbf{G}=(\mathbf{Y}/\sqrt{d})^\top (\mathbf{Y}/\sqrt{d})$. Now, let $\mathbf{Z}$ be a $d\times N$ matrix with columns $\bm{Z}_j$. Define the matrix 
\begin{eqnarray}
	\mathbf{A}:= \frac{\mathbf{Z}^\top \mathbf{Z}}{d}. 
\end{eqnarray}
Let $\lambda_j(\mathbf{A})$ denote the eigenvalue of matrix $\mathbf{A}$. Then, the singular values of $\mathbf{Z}/\sqrt{d}$ are $\sigma_j(\mathbf{Z}/\sqrt{d}) = \sqrt{\lambda_j(\mathbf{A})}$. 
\begin{theorem}(Bai-Yin Theorem)\cite{Bai1993}
	For a $d\times N$ matrix $\mathbf{Z}$ with entries i.i.d $\mathcal{N}(0,1)$ distributed, if  $N / d\rightarrow \beta$, where $\beta$ is a constant in $[0,1]$, then almost surely
	\begin{eqnarray}
		\sigma_{\max}\left(\frac{\mathbf{Z}}{\sqrt{d}}\right) \xrightarrow{a.s.} 1+\sqrt{\beta} \ \ \mathrm{and} \ \ \sigma_{\min}\left(\frac{\mathbf{Z}}{\sqrt{d}}\right) \xrightarrow{a.s.} 1-\sqrt{\beta},
	\end{eqnarray}
	where $\sigma_{\max}$ and $\sigma_{\min}$ are the largest and smallest singular values.
\end{theorem}
In our setting, since $N$ is a constant, then $\beta=0$. Applying Bai-Yin theorem and since almost sure convergence implies convergence in probability, we get
\begin{eqnarray}
	\sigma_{\max}\left(\frac{\mathbf{Z}}{\sqrt{d}}\right)\xrightarrow{P} 1 \ \ \mathrm{and} \ \ \sigma_{\min}\left(\frac{\mathbf{Z}}{\sqrt{d}}\right)\xrightarrow{P} 1, \ \ \mathrm{as} \ d\rightarrow \infty. \label{SingValu}
\end{eqnarray}
Since $\mathbf{A}-\mathbf{I}_N$ is symmetric, its operator norm is
\begin{eqnarray}
	\| \mathbf{A}-\mathbf{I}_N\|_{\mathrm{op}} = \max\left\{\sigma_{\max}^2\left(\frac{\mathbf{Z}}{\sqrt{d}}\right) -1, 1- \sigma_{\min}^2\left(\frac{\mathbf{Z}}{\sqrt{d}}\right) \right\}, \label{op}
\end{eqnarray}
where $\| .\|_{\mathrm{op}}$ denotes the operator norm of the matrix.
By the result in (\ref{SingValu}), we know that both of these values in (\ref{op}) converge to $0$ in probability and hence,
\begin{eqnarray}
	\| \mathbf{A}-\mathbf{I}_N\|_{\mathrm{op}} \xrightarrow{P}0, \ \ \mathrm{as} \ d\rightarrow \infty.
\end{eqnarray} 
Now, for each $j\in\{1,...,N\}$, we know that $\| \bm{Z}_{j}\| ^2 $ is distributed by Chi-squared distribution with $d$ degrees of freedom, i.e., $\| \bm{Z}_j\| \sim \chi_{d}^2$. By results in \cite{Laurent2000} and \cite{Hsu2012}, we have
\begin{eqnarray}
	\mathbb{P}\left[\left| \frac{\| \bm{Z}_j\|^2}{d}-1\right|>\epsilon \right] \leq 2e^{-c_\epsilon d},
\end{eqnarray}
for some $c_\epsilon >0$. Resulting into,
\begin{eqnarray}
	\frac{\| \bm{Z}_j\|^2}{d} \xrightarrow{P} 1.
\end{eqnarray}
Since $N$ is a constant, then
\begin{eqnarray}
	\mathbb{P}\left[\max_{1\leq j\leq N}\left| \frac{\| \bm{Z}_j\|^2}{d}-1\right|>\epsilon\right] \leq 2Ne^{-c_\epsilon d}.
\end{eqnarray}
Hence, it yields again that
\begin{eqnarray}
	\max_{1\leq j\leq N}\left| \frac{\| \bm{Z}_j\|^2}{d}-1\right| \xrightarrow{P}0. \label{P}
\end{eqnarray}
Using the result in (\ref{P}), if we define the diagonal matrix 
\begin{eqnarray}
	\mathbf{D}:= \mathrm{diag}\left(\frac{\| \bm{Z}_1\|}{\sqrt{d}}, ..., \frac{\| \bm{Z}_N\|}{\sqrt{d}}\right),
\end{eqnarray}
then 
\begin{eqnarray}
	\| \mathbf{D} - \mathbf{I}_N\|_{\mathrm{op}}\xrightarrow{P} 0, \ \ \mathrm{as} \ d\rightarrow \infty.
\end{eqnarray}
According to the definition of $\bm{Y}_j=\sqrt{d}\bm{Z}_j/\| \bm{Z}_j\|$, we can write
\begin{eqnarray}
	\mathbf{Y}={\mathbf{Z}}\mathbf{D}^{-1},
\end{eqnarray}
where $\mathbf{D}^{-1}$ is the inverse of matrix $\mathbf{D}$. Therefore, we can write the Gram matrix as
\begin{eqnarray}
	\mathbf{G}=\left(\frac{\mathbf{Y}}{\sqrt{d}}\right)^\top\left(\frac{\mathbf{Y}}{\sqrt{d}}\right) =\left(\mathbf{D}^{-1}\frac{\mathbf{Z}^{\top}}{\sqrt{d}}\right)\left(\frac{\mathbf{Z}}{\sqrt{d}}\mathbf{D}^{-1}\right)= \mathbf{D}^{-1}\mathbf{A}\mathbf{D}^{-1}.\label{G}
\end{eqnarray}

According to (\ref{G}), it is simple to see that
\begin{eqnarray}
	\mathbf{G}-\mathbf{I}_N = \mathbf{D}^{-1}\mathbf{A}\mathbf{D}^{-1} -\mathbf{I}_N= \mathbf{D}^{-1}(\mathbf{A}-\mathbf{I}_N)\mathbf{D}^{-1}+ (\mathbf{D}^{-1})^2 - \mathbf{I}_N.\label{takeop}
\end{eqnarray}
By taking operator norm from both sides of (\ref{takeop}), we have
\begin{eqnarray}
	\| \mathbf{G}-\mathbf{I}_N\|_{\mathrm{op}} \leq \| \mathbf{D}^{-1}\|_{\mathrm{op}}^2 \| \mathbf{A}-\mathbf{I}_N\|_{\mathrm{op}}+\| (\mathbf{D}^{-1})^2-\mathbf{I}_N\|_{\mathrm{op}}.\label{GI}
\end{eqnarray}
We have shown that $\| \mathbf{D}-\mathbf{I}_N\|_{\mathrm{op}} \xrightarrow{P} 0$; therefore, $\| \mathbf{D}^{-1}\|_{\mathrm{op}} \xrightarrow{P} 1$. Because the diagonal entries of $\mathbf{D}$ converges to $1$ in probability; hence, $\| (\mathbf{D}^{-1})^2-\mathbf{I}_N\|_{\mathrm{op}}\xrightarrow{P}0$. We also showed that $\| \mathbf{A}-\mathbf{I}_N\|_{\mathrm{op}} \xrightarrow{P}0$. Combining these results, according to (\ref{GI}), it yields
\begin{eqnarray}
	\| \mathbf{G}-\mathbf{I}_N\|_{\mathrm{op}}\xrightarrow{P} 0 , \ \ \mathrm{as} \ d\rightarrow \infty. \label{fi1}
\end{eqnarray}

For a fixed $\bm{w} \in \mathbb{S}^{d-1}(\sqrt{d})$, the geodesic distance $d_{geo}(\bm{w},\bm{Y}_j)=\sqrt{d}\arccos (\langle \bm{w},\bm{Y}_j \rangle/d)$. Because cosine is decreasing on $[0,\pi]$, then
\begin{eqnarray}
	\cos \frac{r_N(d)}{\sqrt{d}} = \max_{\bm{w}\in \mathbb{S}^{d-1}(\sqrt{d})} \ \min_{1\leq j\leq N}\frac{\langle \bm{w},\bm{Y}_j\rangle}{d}. \label{rN}
\end{eqnarray}
For a fixed $\bm{w}$, the convex linear combination of $\bm{Y}_j$ does not change the minimizer. Hence, we can rewrite the optimization problem in (\ref{rN}) as
\begin{eqnarray}
	\cos \frac{r_N(d)}{\sqrt{d}} = \max_{\bm{w}\in \mathbb{S}^{d-1}(\sqrt{d})} \ \min_{\bm{\gamma}\in \mathcal{X}}\frac{\langle\bm{w},\mathbf{Y}\bm{\gamma}\rangle}{d},
\end{eqnarray}
where 
\begin{eqnarray}
	\mathcal{X}= \left\{\bm{\gamma}=(\gamma_1, ..., \gamma_N) \in [0,1]^N: \sum_{j=1}^{N}\gamma_j = 1\right\}.
\end{eqnarray}
The function $(\bm{w},\bm{\gamma}) \to \langle \bm{w},\mathbf{Y}\bm{\gamma} \rangle$ is bilinear and hence concave-convex. The sets $\mathbb{S}^{d-1}(\sqrt{d})$ and $\mathcal{X}$ are compact and convex. Therefore, by Sion's minimax theorem \cite{Komiya1988}, we can change the order of minimum and maximum as the following
\begin{eqnarray}
	\max_{\bm{w}\in \mathbb{S}^{d-1}(\sqrt{d})}\min_{\bm{\gamma}\in \mathcal{X}}\frac{\langle\bm{w},\mathbf{Y}\bm{\gamma}\rangle}{d} = \min_{\bm{\gamma}\in \mathcal{X}}\max_{\bm{w}\in \mathbb{S}^{d-1}(\sqrt{d})}\frac{\langle\bm{w},\mathbf{Y}\bm{\gamma}\rangle}{d}. \label{minmax}
\end{eqnarray}
Now focusing on the right side of (\ref{minmax}), for a fixed $\bm{\gamma}\in \mathcal{X}$, we can write
\begin{eqnarray}
	\max_{\bm{w}\in \mathbb{S}^{d-1}(\sqrt{d})}\frac{\langle \bm{w},\mathbf{Y}\bm{\gamma}\rangle}{d} = \frac{\| \mathbf{Y}\bm{\gamma}\|}{\sqrt{d}}. 
\end{eqnarray}
Hence, 
\begin{eqnarray}
	\cos \frac{r_N(d)}{\sqrt{d}} = \min_{\bm{\gamma}\in \mathcal{X}}\frac{\| \mathbf{Y}\bm{\gamma}\|}{\sqrt{d}},
\end{eqnarray}
which is equivalent to
\begin{eqnarray}
	\cos^2 \frac{r_N(d)}{\sqrt{d}} = \min_{\bm{\gamma}\in \mathcal{X}} \bm{\gamma}^{\top}\left(\frac{\mathbf{Y}}{\sqrt{d}}\right)^{\top}\left(\frac{\mathbf{Y}}{\sqrt{d}}\right)\bm{\gamma}= \min_{\bm{\gamma}\in \mathcal{X}} \bm{\gamma}^{\top} \mathbf{G}\bm{\gamma}.\label{112}
\end{eqnarray}
Moreover, for all $\bm{\gamma}\in \mathcal{X}$, we have
\begin{eqnarray}
	\left | \bm{\gamma}^{\top}\mathbf{G}\bm{\gamma}-\| \bm{\gamma}\|^2 \right|= \left | \bm{\gamma}^{\top}(\mathbf{G}-\mathbf{I}_N) \bm{\gamma} \right| \leq \| \mathbf{G}-\mathbf{I}_N\|_{\mathrm{op}} \| \bm{\gamma}\|^2 \stackrel{(e)}{\leq}\| \mathbf{G}-\mathbf{I}_N\|_{\mathrm{op}},\label{GGamma}
\end{eqnarray}
where $(e)$ follows from $\bm{\gamma}\in \mathcal{X}$ and hence $\| \bm{\gamma}\|^2 \leq 1$. Therefore, uniformly over all $\bm{\gamma}\in \mathcal{X}$ and for all $\epsilon>0$, by (\ref{GGamma}), we obtain
\begin{eqnarray}
	\mathbb{P}\left[\left | \bm{\gamma}^{\top}\mathbf{G}\bm{\gamma}-\| \bm{\gamma}\|^2 \right|>\epsilon\right] \leq \mathbb{P}\left[\| \mathbf{G}-\mathbf{I}_N\|_{\mathrm{op}} > \epsilon\right]. 
\end{eqnarray}
From (\ref{fi1}), it yields
\begin{eqnarray}
	\bm{\gamma}^\top\mathbf{G}\bm{\gamma} \xrightarrow{P} \| \bm{\gamma}\|^2, \ \ \mathrm{as} \ d\rightarrow \infty.
\end{eqnarray} 
Hence, by (\ref{112}), we have
\begin{eqnarray}
	\cos ^2 \frac{r_N(d)}{\sqrt{d}} \xrightarrow{P} \min_{\bm{\gamma}\in \mathcal{X}}\| \bm{\gamma}\|^2.
\end{eqnarray}
The minimizer of $\| \bm{\gamma}\|^2$ over the simplex $\mathcal{X}$ is $\gamma_1=...=\gamma_N=1/N$, resulting into $\min_{\bm{\gamma}\in \mathcal{X}}\| \bm{\gamma}\|^2 = 1/N$. Therefore, 
\begin{eqnarray}
	\cos \frac{r_N(d)}{\sqrt{d}} \xrightarrow{P} \frac{1}{\sqrt{N}}, \ \ \mathrm{as} \ d\rightarrow \infty. 
\end{eqnarray}
By continuous mapping theorem, we can then write
\begin{eqnarray}
	\frac{r_N(d)}{\sqrt{d}} \xrightarrow{P} \arccos \frac{1}{\sqrt{N}}, \ \ \mathrm{as} \ d\rightarrow \infty.
\end{eqnarray}
By definition of convergence in probability, for any $\epsilon^\prime >0$, we have 
\begin{eqnarray}
	\mathbb{P}\left[\left| \frac{r_N(d)}{\sqrt{d}}-\arccos \frac{1}{\sqrt{N}}\right|>\epsilon^\prime\right]\xrightarrow{P} 0, \ \ \mathrm{as} \ d\rightarrow \infty.\label{Prim2}
\end{eqnarray}

To ensure the upper bound on $\mathbb{P}[\mathcal{B}_{\epsilon}^{c}]$ in (\ref{Prim}) tends to zero as $d\rightarrow \infty$, comparing (\ref{Prim}) with (\ref{Prim2}), we must have
\begin{eqnarray}
	\frac{r^*(d)}{\sqrt{d}}-\arccos u^\prime_\alpha - \epsilon > \epsilon^\prime + \arccos \frac{1}{\sqrt{N}},
\end{eqnarray}
which results into
\begin{eqnarray}
	N < \frac{1}{\cos^2 \left(\frac{r^*(d)}{\sqrt{d}}-\arccos u^\prime_\alpha - \epsilon -\epsilon^\prime\right)}.
\end{eqnarray}
Then, for all sufficiently small $\epsilon$ and $\epsilon^\prime$ and according to the result from Lemma \ref{Lemma1}, we get
\begin{eqnarray}
	N< \frac{1}{\cos^2\left(\sqrt{\frac{2\eta}{\lambda^*}}-\arccos u^\prime_\alpha\right)},
\end{eqnarray}
which completes the proof.
\end{proof}

\section{Proof of Theorem \ref{Thm2}}\label{ProofThm2}
\begin{proof}
By chain rule for mutual information, we have
\begin{eqnarray}
	I(\bm{\Theta}_{1:T};\bm{W}_{1:T}) = \sum_{t=1}^{T}I(\bm{\Theta}_{1:T};\bm{W}_t|\bm{W}_{1:t-1}).\label{chainrule}
\end{eqnarray}
We know that by causality of channels, $\bm{W}_t$ is conditionally independent of $\bm{\Theta}_{t+1:T}$ given $\bm{\Theta}_{1:t}$ and $\bm{W}_{1:t-1}$. Hence, 
\begin{eqnarray}
	I(\bm{\Theta}_{1:T};\bm{W}_t|\bm{W}_{1:t-1}) = I(\bm{\Theta}_{1:t};\bm{W}_t|\bm{W}_{1:t-1}). \label{eq}
\end{eqnarray}
Substituting (\ref{eq}) into (\ref{chainrule}), we get
\begin{eqnarray}
	I(\bm{\Theta}_{1:T};\bm{W}_{1:T}) = \sum_{t=1}^{T}I(\bm{\Theta}_{1:t};\bm{W}_t|\bm{W}_{1:t-1}). 
\end{eqnarray}
For each $t$, we apply the conditional mutual information radius identity \cite{Cai2019} with reference to measure $\sigma$ as
\begin{eqnarray}
	I(\bm{\Theta}_{1:t};\bm{W}_t|\bm{W}_{1:t-1})=\mathbb{E}\left[D\left(P_{\bm{W}_t|\bm{\Theta}_{1:t},\bm{W}_{1:t-1}}\| \sigma\right)\right] - \mathbb{E}\left[D\left(P_{\bm{W}_t|\bm{W}_{1:t-1}} \| \sigma \right)\right],\label{InfoRadius}
\end{eqnarray}
where $P_{\bm{W}_t| \bm{\Theta}_{1:t},\bm{W}_{1:t-1}}$ and $P_{\bm{W}_t|\bm{W}_{1:t-1}}$ are conditional distribution probabilities. The first term in the right hand side of (\ref{InfoRadius}) equals 
\begin{eqnarray}
	\mathbb{E}\left[D\left(P_{\bm{W}_t|\bm{\Theta}_{1:t},\bm{W}_{1:t-1}}\| \sigma\right)\right]=\mathbb{E}\left[D(Q_t(.|\bm{W}_{t-1},\mathcal{K}_{t-1},\bm{\Theta}_t) \| \sigma)\right] \leq \rho_t \label{KLTheotherform}
\end{eqnarray}
by the KL constraint in (\ref{KL}). It remains to lower bound the second term in the right hand side of (\ref{InfoRadius}). Since,
\begin{eqnarray}
	\mathcal{K}_{t}\left(\bm{\theta}_{1:t}\right) \subset \bigcup_{t^\prime=1}^{t}\mathcal{K}_{t^\prime}(\bm{\theta}_{1:t^\prime}),
\end{eqnarray}
then the reliability and non-forgetting constraint implies 
\begin{eqnarray}
	1-\delta_t\leq \mathbb{P}\left[\bm{W}_t\in \mathcal{K}_{t}(\bm{\Theta}_{1:t})\right]\leq \mathbb{P}\left[\bm{W}_t \in \bigcup_{t^\prime=1}^{t} \mathcal{K}_{t^\prime}(\bm{\Theta}_{1:t^\prime})\right]. \label{cupup}
\end{eqnarray}

For each realization $\bm{w}_{1:t-1}$, let
\begin{eqnarray}
	p(\bm{w}_{1:t-1}):=\mathbb{P}\left[\bm{W}_t\in \bigcup_{t^\prime=1}^{t}\mathcal{K}_{t^\prime}(\bm{\Theta}_{1:t^\prime}) \middle \vert \bm{W}_{1:t-1}=\bm{w}_{1:t-1}\right].\label{p}
\end{eqnarray}
By data processing inequality for relative entropy under the measurable map $\bm{w}_t \mapsto \mathbf{1}\left \{ \bm{w}_t \in \bigcup_{t^\prime=1}^{t}\mathcal{K}_{t^\prime}\right \}$, we get
\begin{eqnarray}
	D\left(P_{\bm{W}_t | \bm{W}_{1:t-1}=\bm{w}_{1:t-1}} \| \sigma \right) \geq D_{\mathrm{bin}}\left(p(\bm{w}_{1:t-1}) \| u_t\right).\label{**}
\end{eqnarray}
Taking expectation with respect to $\bm{W}_{1:t-1}$ from both sides of (\ref{**}), and using the convexity of $p \mapsto D_{\mathrm{bin}}(p\| u_t)$, we obtain
\begin{eqnarray}
	&& \mathbb{E}\left[D\left(P_{\bm{W}_t | \bm{W}_{1:t-1}} \| \sigma \right)\right]\geq \mathbb{E}\left[D_{\mathrm{bin}}\left(p(\bm{W}_{1:t-1}) \| u_t\right)\right]\\
	&& \ \ \ \ \ \ \ \ \ \ \ \ \ \ \ \ \ \ \ \ \ \ \ \ \ \ \ \ \ \ \ \ \ \ \ \ \geq D_{\mathrm{bin}}\left(\mathbb{E}\left[p(\bm{W}_{1:t-1})\right] \| u_t\right)\stackrel{(a)}{=}D_{\mathrm{bin}}\left(\mathbb{P}\left[\bm{W}_t\in \bigcup_{t^\prime=1}^{t}\mathcal{K}_{t^\prime}(\bm{\Theta}_{1:t^\prime})\right]\middle |\middle | u_t \right), \label{2circ}
\end{eqnarray}
where $(a)$ follows from (\ref{p}).
Combining (\ref{cupup}) and (\ref{2circ}), we get
\begin{eqnarray}
	\mathbb{E}\left[D\left(P_{\bm{W}_t | \bm{W}_{1:t-1}} \| \sigma \right)\right]\geq d_{+}(1-\delta_t \| u_t):=\inf_{p\in [1-\delta_t, 1]} D_{\mathrm{bin}}(p \| u_t) = \begin{cases}
	D_{\mathrm{bin}}(1-\delta_t \| u_t), & u_t<1-\delta_t\\
	0, & u_t\geq 1-\delta_t
\end{cases}.\label{inf}
\end{eqnarray}
Substituting (\ref{KLTheotherform}) and (\ref{inf}) into (\ref{InfoRadius}) yields
\begin{eqnarray}
	I(\bm{\Theta}_{1:t};\bm{W}_t|\bm{W}_{1:t-1}) \leq \rho_t - d_{+}(1-\delta_t \| u_t).\label{rhominusd}
\end{eqnarray}
Summing (\ref{rhominusd}) over $t$ and using (\ref{chainrule}) implies
\begin{eqnarray}
	I(\bm{\Theta}_{1:T};\bm{W}_{1:T}) \leq \sum_{t=1}^{T} (\rho_t - d_{+}(1-\delta_t \| u_t)).\label{finb}
\end{eqnarray}
We have shown that, for an arbitrary fixed $Q_{1:T}\in \mathcal{Q}(\rho_{1:T}, \delta_{1:T})$, the bound in (\ref{finb}) holds. Therefore, taking the supremum preserves the inequality, which completes the proof.
\end{proof}

\bibliographystyle{IEEEtran}
\bibliography{references_ContinualLearning_Naz_Mirhosseini}

\end{document}